\documentclass{article}
\usepackage{arxiv}
\usepackage[colorinlistoftodos]{todonotes}
\usepackage{soul}
\usepackage{amsmath}

\usepackage[algoruled,linesnumbered]{algorithm2e}
\usepackage{xcolor}
\usepackage{tcolorbox}
\usepackage{amssymb}
\usepackage{amsmath}
\usepackage{multirow}
\usepackage{subcaption}
\usepackage{pdfpages}
\usepackage{booktabs}

\newtheorem{problem}{Problem}

\usepackage[utf8]{inputenc} 
\usepackage[T1]{fontenc}    
\usepackage{hyperref}       
\usepackage{url}            
\usepackage{booktabs}       
\usepackage{amsfonts}       
\usepackage{nicefrac}       
\usepackage{microtype}      
\usepackage{xspace}
\definecolor{myBlue}{RGB}{0,191,255}
\newcommand{\german}{\texttt{german}\xspace}
\newcommand{\credit}{\texttt{credit}\xspace}
\newcommand{\givemecredit}{\texttt{givemecredit}\xspace}
\newcommand{\propublica}{\texttt{propublica}\xspace}

\title{Equalizing Recourse across Groups}

\author{
  Vivek Gupta\thanks{represents equal contribution.}\\
  School of Computing\\
  University of Utah\\
  \texttt{vgupta@cs.utah.edu} \\
   \And
  Pegah Nokhiz$^*$\\
  School of Computing\\
  University of Utah\\
  \texttt{pnokhiz@cs.utah.edu} \\
   \And
  Chitradeep Dutta Roy$^*$ \\
  School of Computing\\
  University of Utah\\
  \texttt{rahduro@cs.utah.edu} \\
   \AND
  Suresh Venkatasubramanian \\
  School of Computing\\
  University of Utah\\
  \texttt{suresh@cs.utah.edu} \\
}

\DeclareUnicodeCharacter{2212}{-}
\DeclareUnicodeCharacter{2032}{-}

\begin{document}

\maketitle
\begin{abstract}
The rise in machine learning-assisted decision-making has led to concerns
about the fairness of the decisions and techniques to
mitigate problems of discrimination. If a negative decision is made about an individual
(denying a loan, rejecting an application for housing, and so on) justice
dictates that we be able to ask how we might change circumstances to get a
favorable decision the next time. Moreover, the ability to change circumstances (a
better education, improved credentials) should not be limited to only those
with access to expensive resources. In other words, \emph{recourse} for
negative decisions should be considered a desirable value that can be
equalized across (demographically defined) groups. This paper describes how to
build models that make accurate predictions while still ensuring that the
penalties for a negative outcome do not disadvantage different groups
disproportionately.

We measure recourse as the distance of an individual from
the decision boundary of a classifier.  We then introduce a regularized
objective to minimize the difference in recourse across groups. We explore
linear settings and further extend recourse to non-linear settings as well as
model-agnostic settings where the exact 
distance from boundary cannot be calculated. Our results show that we can
successfully decrease the unfairness in recourse while maintaining classifier performance.
\footnote{This research was supported in part by the NSF under grant IIS-1633724.}
\end{abstract}

\keywords{Fairness \and Recourse \and Kernels \and Machine Learning}

\section{Introduction}
\label{sec:introduction}

Algorithm-assisted decision making is now ubiquitous in domains as diverse as
employment, education, credit, insurance, and criminal justice. How can we be
sure that the decisions made through such systems will be fair and justifiable?
One way in which we have traditionally placed checks on the power of
decision-making systems, especially in settings where the decisions may have
significant effects on our lives, is by providing the subject of the decision
with a mechanism for recourse \cite{ustun2019actionable}: the ability to ask
how the decision was made and what can be done to change it. This idea of
recourse is a part of the Equal Credit Opportunity Act –an individual can demand
that reasons for a loan denial be provided – and research suggests
that it might also eventually be a component of the right to explanation
implicated in the General Data Protection Regulation of the European Union. But
the idea of recourse is more than a legal concept: it can be viewed as an
ethical and just way to make decision-making more transparent. If a decision
that materially affects us is made using factors that we cannot realistically
change (for example a decision based on our gender or race), then this decision
eliminates our agency and is therefore unjust \cite{rubelalgorithms}. Thus
providing opportunities for recourse is a valuable goal independent of concerns
about fairness and non-discrimination.

Recourse must be feasible to be
useful. If an individual is rejected for a small business loan because of a low
credit score and the recommended “recourse” is to attempt double their net worth, this
is not a reasonable outcome.  The situation is even worse if individuals from a
different demographic are given a different recommendation – for example, to
take an (inexpensive) course on money management for small businesses – that is
easier to act on. In other words, providing an explanation is useful, but
providing paths for recourse that are feasible and do not unduly burden
disadvantaged groups is crucial to provide transparent opportunities for
advancement. 

\paragraph{Our Work.}

In this paper, we build classifiers that maintain prediction accuracy
while also ensuring opportunities for feasible recourse across groups. To do
this, we define a general notion of recourse as the
distance from the decision boundary (inspired by the work of
\cite{ustun2019actionable}) and utilize it as a regularizer. 
In addition, we extend previous linear approaches
for recourse \cite{ustun2019actionable} to non-linear settings, including both
SVMs as well as models defined by black-box classifiers where we cannot
explicitly compute the distance to the decision boundary. 
Our results show that we can
successfully equalize recourse across groups in both SVM and model agnostic
settings while maintaining accuracy. 

\textbf{On the ethical validity of equalized recourse.} Recourse as a good is different from fairness/nondiscrimination. Regardless of the decision process, recourse captures the ethical notion of agency: that individuals should be able take actions to rectify their circumstances, and that decisions are not based on immutable factors. Recourse might not be cheap – but it should not be intrinsically infeasible. The idea of equalizing recourse across groups is then more than “if it’s good it should be good for all”. Rather, if we like recourse we don’t want a system to provide token recourse to one group whilst covertly discriminating. That is the justification for our work. We don’t claim that recourse must be universally applicable. Rather, it focuses on a specific kind of unfairness and seeks to address it when appropriate. We also strenuously push back on the idea that equalizing recourse (or even other fairness measures) “lowers the bar for the disadvantaged”. This perspective assumes that the bar was set correctly to begin with, rather than potentially being evidence of structural disadvantage.

\section{Related work}
\label{Related}

The idea of recourse as a desirable property of a classifier was introduced and
formalized by \cite{ustun2019actionable}. In their work they analyze a linear
classifier and show how to use integer linear programming to generate a set of
actionable input variables (flipsets) to change the prediction for any
individual. Their work can be thought of as part of a larger effort in fairness
via \emph{counterfactual analysis} \cite{wachter2017counterfactual} where the goal was to define
fairness in terms of counterfactual properties of classifications.  Strategic
classification techniques \cite{hardt2016strategic,hu2019disparate, milli2019social} which aim
to modify the input in order to trick the classifier and achieve the desired
decision are another concept related to recourse.

Our work fits within the larger setting of methods that try to find classifiers
that satisfy some notion of fairness via constraints during the model learning
process such as for demographic parity, equalized odds
\cite{zafar2017fairness,zafar2017fairness2} and individual fairness
\cite{dwork2012fairness}. But recourse is orthogonal to those notions of fairness being studied. That is, it makes sense to talk about recourse even if the classification is unfair in other respects, because measures of fairness apply at the point of decision, whereas recourse is a post-decision analysis. Equalizing recourse affects distance to the boundary, but does not necessarily affect the decision itself. Having said that, studying the interaction between fair classification and recourse is a worthwhile topic for further exploration. Our approach in this paper involves shifting a baseline classifier
through (iterated) processing which is similar to works such as
\cite{fish2016confidence,goh2016satisfying,dwork2018decoupled,pleiss2017fairness,woodworth2017learning}. Recourse can be viewed as a kind of explanation: an indicator of how to change
one's classification. We employ the well-known method
LIME \cite{ribeiro2016should,ribeiro2018anchors} to construct local explanations
that guide our search for a classifier that equalizes recourse. 
\cite{dhurandhar2018explanations} recently proposed a method to
provide minimal absent feature sets from examples which are sufficient for
flipping their prediction.

\section{Definitions}
\label{sec:Fairness}

Recourse (as defined by \cite{ustun2019actionable}) is the cost of changing a
decision from negative to positive. Formally, assume that we are given a
classifier $h: \mathbb{R}^d \to \{-1,1\}$ and a metric $d: \mathbb{R}^d \times \mathbb{R}^d
\to \mathbb{R}$. The distance metric $d$ captures the ``cost'' of changing the features of a point
$\mathbf{x}$, i.e., the effort involved in moving from $\mathbf{x}$ to
$\mathbf{x}'$ is $d(\mathbf{x}, \mathbf{x}')$.

Given a point $\mathbf{x}$ such that $h(\mathbf{x}) = -1$, the
\emph{recourse} of $\mathbf{x}$ is defined as $r(\mathbf{x}) = \min_{h(\mathbf{x}') = 1} d(\mathbf{x}, \mathbf{x}')$.
In other words, the recourse of a point is the (closest) distance from that point to the
classifier.

The recourse of a set $S$ is the average recourse of all points in $S$. Given a
set of labeled pairs $(\mathbf{x}_i, y_i)_{i = 1\ldots n} \in \mathbb{R}^d \times
\{-1,1\}$ with a \emph{group membership} function $g : \mathbb{R}^d \to \{-1,1\}$,
let $G_j = \{ \mathbf{x}_i \mid g(\mathbf{x}_i) = j\}$.

\begin{problem}[Equalizing Recourse]
  Given a set of labeled pairs $(\mathbf{x}_i, y_i)_{i = 1\ldots n} \in \mathbb{R}^d \times
\{-1,1\}$ and $\epsilon > 0$, find a classifier $h$ such that the loss $ L = \sum_i
\ell(h(\mathbf{x}_i), y_i)$ is minimized under the constraint that
$u^+ = |r(G^-_1) - r(G^-_{-1})| \le \epsilon$
where $G^-_i = \{ \mathbf{x} \in G_i \mid h(\mathbf{x}) = -1\}$.
\label{thm:problem1}
\end{problem}

\paragraph{Note.}

While we borrow the definition of recourse from \cite{ustun2019actionable}, we
do not also adopt their notion of a set of permissible actions to achieve
recourse. This is motivated by the fact that we are trying to optimize recourse
across groups rather than generate a flip set for a specific individual. 

\section{Optimizing for recourse}
\label{sec:optimizing-recourse}

We now present our optimization frameworks for solving
Problem~\ref{thm:problem1}. We start with the case where the classifier can
be written explicitly.

\subsection{Optimizing for recourse with an explicit classifier}

For a classifier with a clear functional expression  $h:X\to Y$, we can write down
expressions for the minimum distance of a point $x$ to the boundary of the
classifier that is recourse $r(x)$. We then can simultaneously optimize the loss
as well as the recourse. 

Below, we show the optimization for equal recourse in kernelizable support vector machines; however, our method is quite general and can be transferred to other kernelizable classifiers such as Perceptron.

\textbf{Support Vector Machines:} Consider the case of support vector machines. For any linear classifier of the form $h(\mathbf{x}) = \mathbf{w}^\top\mathbf{x} + b$, the distance to the decision boundary from a point $x$ is $\frac{\mathbf{w}^\top
\mathbf{x} + b}{\|\mathbf{w}\|}$. In general, different features might be harder to
modify, which can be modeled by scaling each dimension $j$ with a factor
$c_j$. Effectively this amounts to applying a linear transformation $\mathbf{x}
\to \mathbf{C}\mathbf{x}$ via a
diagonal matrix  $\mathbf{C}$ where $C_{jj} = c_j$, and then computing the
corresponding distance $\frac{\mathbf{w}^\top
  \mathbf{C}\mathbf{x} + b}{\|\mathbf{w}\|}$.
If we now introduce a kernel $\mathbf{K(\cdot, \cdot)}$, the above distance calculation
must happen in the lifted \emph{feature space} where $\mathbf{C}\mathbf{x}$ is
mapped to $\Phi(\mathbf{C}\mathbf{x})$, and so the corresponding distance
to the boundary is given by

\[ r(\mathbf{x}) = \frac{\mathbf{w}^\top
  \Phi(\mathbf{C}\mathbf{x}) + b}{\|\mathbf{w}\|} \]

The recourse of a set $(G^-_i)$ is merely the average recourse of the points in
the set and the recourse difference  $u^+ = |r(G^-_1) - r(G^-_{-1})|$.

\textbf{SVM primal form.} We formulate a data-driven primal-form constraint of kernel SVM inspired from \cite {darkner2012data} with an explicit recourse constraint.
\begin{align*}
    & \text{min}_{\mathbf{w}, b, \mathcal{E}, \epsilon}
    & &  \frac{1}{2}\mathbf{w}^T\mathbf{w} + \lambda\mathcal{E} + \nu\sum_{i=1}^n\epsilon_i\\
    & \text{such that} & & -\mathcal{E} \leq u \leq \mathcal{E} \\
    & & & y_i(\mathbf{w}^T\mathbf{\mathbf{x_i}} + b) \geq 1 - \epsilon_i \text{ and } \epsilon_i  \geq 0 \qquad \qquad\forall i \in [1\ldots n] \\
    & \text{where}  & &   u  = r(G^-_1) - r(G^-_{-1})
\end{align*}

$\mathcal{E}$ specifies a bound on recourse difference and $\lambda$ controls the relative
weight of $\mathcal{E}$ in the optimization. $\nu , \epsilon$ are slack variables and $\epsilon$ is the slack for SVM classification. 

\textbf{SVM dual form.} The dual form of the kernel SVM with implicit recourse constraints is the following:

\begin{align}
L(\mathbf{w}, b, \mathcal{E}, \epsilon, \rho_1, \rho_2, \gamma, \delta) &= \frac{1}{2}\mathbf{w}^T\mathbf{w} + \lambda\mathcal{E} + \nu\sum_{i=1}^n\epsilon_i + \rho_1(u-\mathcal{E}) -\rho_2(u + \mathcal{E})\nonumber\\
&- \sum_{i=1}^n\gamma_i(y_i(\mathbf{w}^T\mathbf{x_i} + b) + \epsilon_i -1) - \sum_{i=1}^n\delta_i\epsilon_i 
\label{eq:lagrangian}
\end{align}  

where $\rho_1$, $\rho_2$, $\gamma$, and $\delta$ are dual variables.

We use the $KKT$ conditions with respect to  $\mathbf{w}, b, \mathcal{E},$ and $\epsilon$ (equating gradient to 0) to find the optimal parameters.
\begin{align}
\frac{\partial{L}}{\partial{\mathbf{w}}} &= \mathbf{w} + \rho_1\frac{\partial{u}}{\partial{\mathbf{w}}} - \rho_2\frac{\partial{u}}{\partial{\mathbf{w}}} - \sum_{i=1}^n\gamma_i\mathbf{x_i}y_i = 0\label{eq:Lwl}\\
\frac{\partial{L}}{\partial{b}} &=\rho_1\frac{\partial{u}}{\partial{b}} - \rho_2\frac{\partial{u}}{\partial{b}} - \sum_{i=1}^n\gamma_iy_i = 0\label{eq:Lb}\\
\frac{\partial{L}}{\partial{\mathcal{E}}} &= \lambda - \rho_1 - \rho_2 = 0\label{eq:LE}\\
\frac{\partial{L}}{\partial{\mathcal{\epsilon}_i}} &=\nu - \gamma_i - \delta_i = 0\label{eq:Leps}
\end{align}

\textit{Iterative Procedure:} Since recourse is dependent on the prediction labels, that is which example is classified as negative, to calculate recourse for next iteration ($t+1$), we can use the predictions of the last iteration($t$) of SVM ($y^t = h^t(x)$) when defining $u$. Then, we can re-write $u$ as the following for the optimization,
\begin{align}
u &=\sum_{i=1}^n g(\mathbf{x_i}) \frac{\left(1-h^t(\mathbf{x_i})\right)}{2|G_{g(\mathbf{x_i})}|} (\mathbf{w}^T\mathbf{C}\mathbf{x_i} + b)\label{eq:recoursesimplifypartial}
\end{align}
where, $h^t(\mathbf{x_i})$ $\in$ $\{1, -1\}$ denote the classification of the $\mathbf{x_i}$ sample at iteration $t$, and $|G_{g(\mathbf{x_i})}|$ is the cardinality of set $G_{g(\mathbf{x_i})}$ to which $\mathbf{x_i}$ belongs. 

\textit{Recourse Scaling: } We scale the recourse by $\|\mathbf{w}\|$, similar to \cite{darkner2012data}, because we care mostly about the relative difference between recourse. moving across the boundary requires traveling at least $\frac{1}{2\|\mathbf{w}\|}$ and so we can measure the difference in recourse in units of the margin by dividing by the margin $\frac{1}{\|\mathbf{w}\|}$ which is equivalent to multiplying by $\|\mathbf{w}\|$. 

Let's denote $p_i$ = $\frac{\left(1-h^t(\mathbf{x_i})\right)}{2{|G_{g(\mathbf{x_i})}|}}$, C as the diagonal cost matrix, by substituting in equations \eqref{eq:recoursesimplifypartial}, \eqref{eq:Lwl} and \eqref{eq:Lb},
\begin{align}
u &=\sum_{i=1}^n g(\mathbf{x_i})p_i (\mathbf{w}^T\mathbf{C}\mathbf{x_i} + b)\\
\frac{\partial{L}}{\partial{\mathbf{w}}} &=\mathbf{w} + \sum_{i=1}^n\left(g(\mathbf{x_i})p_i(\rho_1-\rho_2)\mathbf{C}\mathbf{x_i} -\gamma_iy_i\mathbf{x_i}\right) = 0\label{eq:NLwl} \\
\frac{\partial{L}}{\partial{b}} &= \sum_{i=1}^n\left(g(\mathbf{x_i})p_i(\rho_1-\rho_2) -\gamma_iy_i\right) = 0\label{eq:NLb}
\end{align}

Equating equations \ref{eq:LE} and \ref{eq:Leps} to zero cancels the terms involving $\mathcal{E}$ and $\epsilon_i$ in the Lagrangian (equation \ref{eq:lagrangian}).
\begin{align}
L &= \frac{1}{2}\mathbf{w}^T\mathbf{w} + \rho_1u -\rho_2u -\sum_{i=1}^n\gamma_iy_i(\mathbf{w}^T\mathbf{x_i} + b) + \sum_{i=1}^n\gamma_i\label{eq:SL1}
\end{align}

We now simplify the terms in the Lagrangian to obtain \ref{eq:SL1} by equating equations \ref{eq:NLwl}, \ref{eq:NLb} to zero and substituting $\mathbf{w}$. 
\begin{align}
L &= \frac{1}{2}\left[\sum_{i=1}^n\left(\gamma_iy_i\mathbf{x_i}-g(\mathbf{x_i})p_i(\rho_1-\rho_2)\mathbf{C}\mathbf{x_i}\right)\sum_{j=1}^n\left(\gamma_jy_j\mathbf{x_j}-g(\mathbf{x_j})p_j(\rho_1-\rho_2)\mathbf{C}\mathbf{x_j}\right) \right]\nonumber\\
&+\sum_{i=1}^ng(\mathbf{x_i})p_i(\rho_1 -\rho_2)\left((\mathbf{C}\mathbf{x_i})^T\sum_{j=1}^n\left(\gamma_jy_j\mathbf{x_j}-g(\mathbf{x_j})p_j(\rho_1-\rho_2)\mathbf{C}\mathbf{x_j}\right) + b\right)\nonumber\\ 
&-\sum_{i=1}^n\gamma_iy_i\left(\left(\sum_{j=1}^n(\gamma_jy_j\mathbf{x_j}-g(\mathbf{x_j})p_j(\rho_1-\rho_2)\mathbf{C}\mathbf{x_j}\right)^T\mathbf{x_i} + b\right) + \sum_{i=1}^n\gamma_i\label{eq:SL2}
\end{align}

\noindent Substitute $K_i$ $=$ $g(\mathbf{x_i})p_i(\rho_1-\rho_2)$ and $H_i$ $=$ $\gamma_iy_i$, we get the following
\begin{align}
L &= \frac{1}{2}\left[\sum_{i=1}^n\left(H_i\mathbf{x_i}-K_i\mathbf{C}\mathbf{x_i}\right)\sum_{j=1}^n\left(H_j\mathbf{x_j}-K_j\mathbf{C}\mathbf{x_j}\right) \right] \nonumber + \sum_{i=1}^n\left(K_i(\mathbf{C}\mathbf{x_i})^T\sum_{j=1}^n\left(H_j\mathbf{x_j}-K_j\mathbf{C}\mathbf{x_j}\right)+ K_ib\right)\nonumber\\
&-\sum_{i=1}^nH_i\left(\left(\sum_{j=1}^n(H_j\mathbf{x_j}-K_j\mathbf{C}\mathbf{x_j}\right)^T\mathbf{x_i} + b\right) + \sum_{i=1}^n\gamma_i
\end{align}
Upon further simplification and cancelling the $b$ terms using equation $\ref{eq:NLb}$, we obtain the following:
\begin{align}
L &= \sum_{i=1}^n\sum_{j=1}^nH_iK_j(\mathbf{C}\mathbf{x_i})^T\mathbf{x_j}-\frac{1}{2}\left[\sum_{i=1}^n\sum_{j=1}^nH_iH_j\mathbf{x_i}^T\mathbf{x_j}+\sum_{i=1}^n\sum_{j=1}^nK_iK_j(\mathbf{C}\mathbf{x_i})^T\mathbf{C}\mathbf{x_j}\right] \nonumber \\ &+ \sum_{i=1}^n\gamma_i
\end{align}

The final Lagrangian formulation after simplification would be:
\begin{align}
        & \text{min}_{\rho_1, \rho_2, \gamma_i}
        & &  \sum_{i=1}^n\sum_{j=1}^n\gamma_i(\rho_1-\rho_2)g(\mathbf{x_j})p_jy_i(\mathbf{C}\mathbf{x_j})^T\mathbf{x_i}\nonumber\\
        & & &-\frac{1}{2}\left[\sum_{i=1}^n\sum_{j=1}^n\gamma_i\gamma_jy_iy_j\mathbf{x_i}^T\mathbf{x_j}+\sum_{i=1}^n\sum_{j=1}^ng(\mathbf{x_i})g(\mathbf{x_j})p_ip_j(\rho_1-\rho_2)^2(\mathbf{C}\mathbf{x_i})^T\mathbf{C}\mathbf{x_j}\right] \nonumber \\
        & & &+ \sum_{i=1}^n\gamma_i\\
        & \text{s.t} & & 0 \leq \rho_1,\rho_2 \leq \lambda \quad ; \quad \rho_1 + \rho_2 = \lambda\nonumber\\
        & & & 0 \leq \gamma_i \leq  \nu \qquad \qquad \forall i \in [1\ldots n] \nonumber \\ & & & \sum_i^n (g(\mathbf{x_i})p_i(\rho_1-\rho_2) -\gamma_iy_i)=0 \nonumber
        \label{eq:dual}
\end{align}

We denote $p_i$ = $\frac{\left(1-h^t(\mathbf{x_i})\right)}{2{|G_{g(\mathbf{x_i})}|}}$, where $h^t(\mathbf{x_i})$ $\in$ $\{1, -1\}$ is the classification of the $\mathbf{x_i}$ sample at iteration $t$, and $|G_{g(\mathbf{x_i})}|$ is the cardinality of set $G_{g(\mathbf{x_i})}$ to which $\mathbf{x_i}$ belongs. Here, $C$ is the diagonal cost matrix.

To write the final Quadratic formulation, we need to introduce a pseudo recourse point ($x_{n+1},y_{n+1}$), where $\mathbf{x_{n+1}} = \sum_{j=1}^{n}g(\mathbf{x_j})p_j\mathbf{C}\mathbf{x_j}$ and $y_{n+1} = 1$, where, $p_i$ = $\frac{\left(1-\textnormal{sgn}(h^t(\mathbf{x_i}))\right)}{2{|G_{g(\mathbf{x_i})}|}}$, where $|G_{g(\mathbf{x_i})}|$ is the cardinality of the set $G_{g(\mathbf{x_i})}$ to which $\mathbf{x_i}$ belongs. Let's denote a new dual variable $\gamma_{n+1} = \rho_2 - \rho_1$ and new constraint constant $y^c = \sum_{j=1}^{n}g(\mathbf{x_j})p_j$.

\noindent Simplified Lagrangian dual formulation with the pseudo recourse point will be:
\begin{align}
        & \text{min}_{\gamma_i}\quad -\frac{1}{2}\sum_{i=1}^{n+1}\sum_{j=1}^{n+1}\gamma_i\gamma_jy_iy_j\mathbf{x_i}^T\mathbf{x_j} +\sum_{i=1}^n\gamma_i\nonumber\\
        & \text{s.t} \quad \quad -\lambda \leq \gamma_{n+1} \leq \lambda\\
        & \quad \quad \quad \quad 0 \leq \gamma_i \leq  \nu \qquad \qquad \forall i \in [1\ldots n] \nonumber \\ 
        & \quad \quad \quad \quad \sum_i^{n} \gamma_iy_i + \gamma_{n+1}y^c=0 \nonumber
\end{align}

\noindent The final Quadratic form (QuadForm) for dual SVM is :
\begin{align}
 & \text{min}_{\mathbf{\mu}} \quad \frac{1}{2} \mathbf{\mu}^T \mathbf{M} \mathbf{\mu} + \mathbf{e}^T \mathbf{\mathbf{\mu}}\nonumber \\
 & s.t. \quad  \mathbf{A}\mathbf{\mu} = \mathbf{0} ; \quad \mathbf{I^-}\mathbf{\mu} \leq \mathbf{a} ; \quad \mathbf{I^+}\mathbf{\mu} \leq  \mathbf{b}
 \label{svm:quadform}
\end{align}

Here, $\mathbf{M}$ has entries $\mathbf{M}_{(i\leq n,j \leq n)} = y_i y_j \mathbf{K(x_i,x_j)}$, $\mathbf{M}_{(n+1,n+1)} = \sum_{(i,j = 1)}^{n} g(\mathbf{x_i})g(\mathbf{x_j}) p_i p_j \mathbf{K(Cx_i,Cx_j)}$,  and $\mathbf{M}_{(i \leq n,n+1)} = \sum_{(j = 1)}^{n} g(\mathbf{x_j}) y_i p_j \mathbf{K(x_i,Cx_j)}$,  where 
$\mathbf{K(x_i,x_j)} = \langle \Phi(\mathbf{x_i}) \cdot \Phi(\mathbf{x_j}) \rangle$, $\mathbf{K(x_i,Cx_j)} = \langle \Phi(\mathbf{x_i}) \cdot \Phi(\mathbf{Cx_j}) \rangle$, and $\mathbf{K(Cx_i,Cx_j)} = \langle \Phi(\mathbf{Cx_i}) \cdot \Phi(\mathbf{Cx_j}) \rangle$ $\forall$ $\{i,j\}$ $\in$ $\{1, \dots, n\}$. Note that $\mathbf{M}$ is a symmetric and positive semi definite matrix.

Here, we have variables $\mathbf{\mu} = [\gamma_1 \gamma_2 \dots \gamma_{n+1}]$, equality matrix $\mathbf{A} = [y_1 \ldots y_n, y^c]$, where $y^c = \sum_{j=1}^{n}g(\mathbf{x_j})p_j$ inequality matrices  $\mathbf{I^-} = -\mathbf{I}_{(n+1)}$ and $\mathbf{I^+} = \mathbf{I}_{(n+1)}$, where  $\mathbf{I}_{(n+1)}$ is the identity matrix, $\mathbf{a} = [0_n \lambda]$, $\mathbf{b} = [\nu_n \lambda]$ and $\mathbf{e} = [-1_n 0]$. 

This form is compatible with any quadratic programming solver (CVXOPT) \cite{andersen2013cvxopt} which results in the optimal values of $\mathbf{\mu}^*$, i.e., the optimal values of $\gamma^*$. Sometimes, solving a complete QP is intractable as it take an order $n^3$. However, we can solve sub-problem to speed out the QP. We can use the standard SMO technique \cite{zeng2008fast,keerthi2001improvements} which optimise the dual variable's efficiently.

\noindent We can calculate $\mathbf{w}$ by equating \ref{eq:NLwl} to $0$ and substituting $\gamma_{n+1}^* = \rho_2-\rho_1$ :

\begin{equation}
\mathbf{w^*} = \sum_{i=1}^n\left(g(\mathbf{x_i})p_i\gamma_{n+1}^*\mathbf{C}\mathbf{x_i} + \gamma_i y_i\mathbf{x_i}\right)\label{eqn:wfinalform}
\end{equation}

For a linear kernel $\mathbf{w}$ will be a vector expressed in a closed form. However, for any other kernel, we cannot calculate $\mathbf{w}^*$ explicitly because of the kernel; however, we can explicitly calculate $\|\mathbf{w}^*\|$ as follows because of kernel property (obtained by applying $\mathbf{w}^T \mathbf{w}$ and some simplification):

\begin{equation}
    \|\mathbf{w}^*\| = \mathbf{\mu}^{*T} \mathbf{M} \mathbf{\mu}^* \label{eqn:normwfinal}
\end{equation}

\noindent For calculating the bias $b^*$, we need to use the support vectors, and those samples whose $\gamma_i^*$ is significantly above a certain threshold $t$ should be on margins. To calculate the final $b^*$ value, we can take means of all values from support vectors:
\begin{equation}
    b_i = y_i - \mathbf{w^*}^T\mathbf{x_i}
\end{equation}
\begin{equation}
    b_i = y_i - \sum_{j=1}^n\left(g(\mathbf{x_j})p_j\gamma_{n+1}^*\mathbf{K(\mathbf{C}\mathbf{x_j}},\mathbf{x_i}) + \gamma_i y_j \mathbf{K(\mathbf{x_j},\mathbf{x_i}})\right)
\end{equation}
\begin{equation}
    b^{*} = \frac{\sum_{i=1}^{\#suppvec}b_i}{\#suppvec}
\end{equation}

The final classifier decision for a new prediction example $x_p$ can be obtained as follows:

\begin{equation}
 h(x_{p}) =  \textnormal{sgn}(\mathbf{w}^T\mathbf{x_{p}} + b^*)
 \label{eqn:prediction}
\end{equation}

Substitute $\mathbf{w}$ from \ref{eqn:wfinalform} into \ref{eqn:prediction} yielded the final expression,

\begin{equation}
 h(x_{p}) =  \textnormal{sgn}(\sum_{i=1}^{n}\gamma_i^*y_i\mathbf{K(x_i,x_p))} + \gamma_{n+1}^*\sum_{i=1}^{n} g(\mathbf{x_i}) p_i \mathbf{K(Cx_i,x_p)} + b^*)\label{eqn:finaldecisionexpression}
\end{equation}

We can similarly calculate the final recourse ($u^+$)) by substituting $\mathbf{w}$ from \ref{eqn:wfinalform} into \ref{eq:recoursesimplifypartial} as follows,

\begin{align}
\Big| \frac{\sum_{j=1}^n g(\mathbf{x_j}) \frac{\left(1-\textnormal{sgn}(h(\mathbf{x_j}))\right)}{2{|G_{g(\mathbf{x_j})}|}} \left(\sum_{i=1}^{n}\gamma_i^*y_i\mathbf{K(x_i,Cx_j))} + \gamma_{n+1}^*\sum_{i=1}^{n} g(\mathbf{x_i}) p_i \mathbf{K(Cx_i,Cx_j)} + b^*\right)}{\|\mathbf{w}^*\|}\Big|\label{eq:recoursesimplify}
\end{align}

where, $\|\mathbf{w}^*\| = \mathbf{\mu}^{*T} \mathbf{M} \mathbf{\mu}^*$ from \ref{eqn:normwfinal} , and the rest of terms are as defined before.

\noindent Here, in equations \ref{eqn:finaldecisionexpression} and \ref{eq:recoursesimplify} we can use any suitable kernel, e.g., polynomial, linear, radial etc. to obtain the kernel matrix terms.

\paragraph{An iterative procedure}

While the above formulation allows us to equalize recourse, it cannot be solved as written. This is because the specification of the dual requires the knowledge of which points are classified negatively due to the calculation of $h(x_j)$. However, for any \emph{fixed} classifier ($h(x)$), we can identify the points that are classified negatively and use that to compute a new recourse-equalized classifier. This suggests an iterative strategy where the classifier from the previous round (t) i.e. $h^t(x)$ is used to identify negatively-classified points which are then used to compute the
classifier for this round (t+1) i.e. ($h^{t+1}(x)$). We summarize this in Algorithm~\ref{algo:PPA}. 

\paragraph{An example.}

\begin{figure}[htbp]   
    \centering
    \scalebox{.75}{
    \includegraphics[scale=0.44]{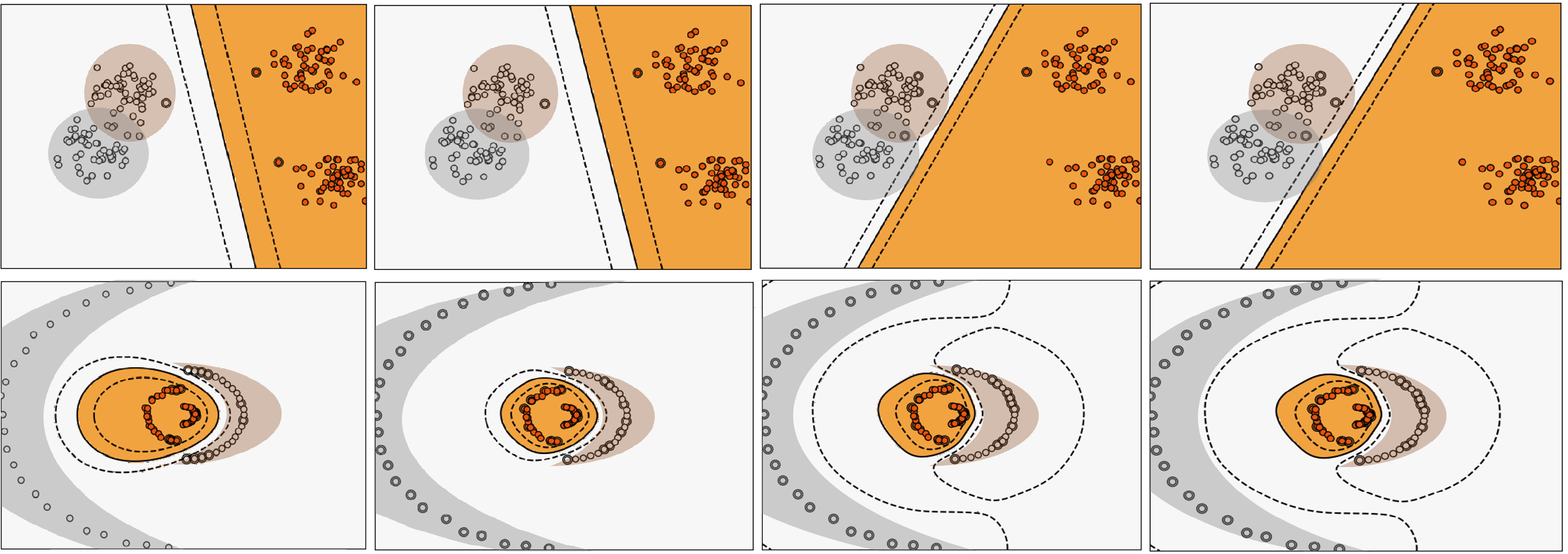}}
    \caption{Equalizing recourse across groups, from left (vanilla SVM ) to right ($\lambda$ increased).}
    \label{fig:all}
    \vspace{-1.0em}
\end{figure}

We applied Algorithm \ref{algo:PPA} to two synthetic toy datasets to analyze our approach. In  \ref{fig:all}, we can see the motivation behind introducing recourse as a regularizer to be equalized across groups. We illustrate the positive class on the solid orange region with red datapoints. Conversely, the negative class is shown on the solid white background. The decision boundary is a solid black line that distinguishes the two areas. The dashed lines are the indicators of the margins. Our goal is to equalize recourse for the negatively classified groups $G^-_{1}$ in the brown transparent region that is closer to the boundary and the farther $G^-_{-1}$ in the gray transparent region which has a larger recourse. As we increase $\lambda$ and thus the importance of equalizing recourse, we observe that in the linear case, the decision boundary tilts such that the under-represented negative group will get closer to the decision boundary and the recourse is equalized. In the non-linear case, the boundary margins (dashed lines) are uni-modal around the positive class (brown), thus making it farther from the outer negative class (gray). As we increase $\lambda$, the boundary margins (dashed lines) become bi-modal with two dashed margins around both groups which means recourse is equalized.   

\textbf{Flipset generation.} Similar to \cite{ustun2019actionable}, we can generate flipsets by finding the closest points to the classifier boundary according to our notion of recourse. This is similar to the linear programming optimization approach proposed by \cite{ustun2019actionable}. One difference is since our attributes are continuous, we have more freedom in finding the flipsets since we know the exact closest boundary points according to our recourse definition. Besides, if we have constraints on attributes (e.g., some features are immutable and cannot be changed such as gender) we can use the ILP/LP idea used in \cite{ustun2019actionable} and utilize it in our balanced recourse classifier to obtain the required flipsets. Thus, our approach is orthogonal to \cite{ustun2019actionable}'s ILP/LP and can generate flipsets. 

One might think of extending the ILP-based formulation \cite{ustun2019actionable} for obtaining fairness in recourse on top of generating flipsets without changing the boundary. For example, to achieve equal recourse one could optimize $\min |\sum_{x_i \in G^-_{1}} c_i{v_i} - \sum_{x_j \in G^-_{-1}} c_j{v_j}|$ using linear programming for finding flip-sets which ensure equal recourse. Here, $c_iv_i$ is the cost of changing features of $x_i$ to achieve recourse using the action set $v_i$ as defined in \cite{ustun2019actionable}. However, this idea contradicts the original notion of minimum actionable flipsets, which is achieved by minimizing $\min |\sum_{x_i \in G^-_{\{1,-1\}}} c_i{v_i}|$. The flipset solution for equal recourse might not necessarily be of the minimum cost. So there might be other low-cost flipsets for actionable recourse. Thus, the individuals always have the right to choose the other flipsets that have a lower cost for them compared to the ones that remove the unfairness. Hence, changing the boundary is essential for yielding equal recourse across groups. In our case, we are changing the boundary to ensure that the minimum flipsets yields equal recourse.

\subsection{Model agnostic settings}

The previous approach for equalizing recourse is only applicable where we have an explicit mathematical formulation of the distance of a point from a classifier boundary and also we know the full objective of the underlying model. This is not the case with any arbitrary non-linear classifier. Also this scheme is not feasible in the model agnostic setting with black-box access. Moreover, finding a closed-form formula of the minimum distance of a point from a non-linear boundary (hence recourse) is either hard or infeasible in general. 

To remedy this situation, we propose an approach that is based on LIME, a technique introduced by \cite{ribeiro2016should} to provide a local explanation for black-box classifiers. LIME works by sampling points from a normal distribution centered at a particular point or the mean of the dataset. LIME then fits a linear model using weighted regression for the sampled points against the decision produced by a black-box classifier. Our method hinges on the idea that LIME lets us approximate a complex decision boundary by training a linear classifier for every individual datapoint, and therefore, allowing us to compute the approximate distance of a point from the actual non-linear decision boundary. Once we have distances (and therefore the estimates for recourse $r(G_i)$ for each group), we can build a new classifier. However, since the classifier is a black-box we cannot explicitly introduce a regularization term as before. 

Therefore, to change the black box non-linear boundary to equalize recourse, we propose an example-dependent weighted learning approach. The idea is that we can retrain the black-box classifier by passing the weight parameters for each sample point. Intuitively, we will re-weight points so that groups with large recourse (i.e., a larger distance to the decision boundary) are given a lower weight than groups with small recourse. The effect of this is to move the decision boundary \emph{away} from the highly weighted points and towards the lower-weighted points, which is the desired effect.

\paragraph{Algorithm.} At first, we generate a set of samples (neighborhood) using multivariate normal centered at the mean of the dataset and scaled accordingly. We then train the black-box classifier. For every datapoint in the training set 10 most important features are chosen using a weighted regression using an exponential kernel with Euclidean distance metric for the sample weights. Now finally a ridge linear classifier is trained on the samples using only the selected features. This classifier is used to calculate the approximate distance of the datapoint from the original black-box classifiers boundary. We do this for all the training points from the trained classifier's boundary. And we average it over several sets of samples normalized by the difference between maximum and minimum distances for the negative class. We keep the weights of points in the positive class ($h(x) = 1$),
unchanged (i.e. 1). To calculate the weight of point ($x_i$) in the negative class ($h(x_i) = -1$), we estimate the normalized minimum average distance
(namd($x_i$)) of $x_i$ from the decision boundary \vspace{-0.5em}
\begin{equation*}
    \text{namd}{(x_i)} = \frac{\min_{x_l \in X^-}(dist^{AVG}(x_l))}{dist^{AVG}(x_i)}
\end{equation*}

Here $X^-$ is set of negatively classified points, We can set the weight $sw(x_k)$ of any $x_k$ point by using the following formula,
\vspace{-1em}
\begin{align}
sw(x_k) = \begin{cases}
1 & \text{$h(x_k) = 1$}\\
namd(x_k)  & \text{otherwise}
\label{eqn:weightcalculation}
\end{cases}
\end{align}

Now, we retrain the classifier with these new sample weights. The classifier tries to be more certain in predicting the class of a point which is weighed more than the ones that are weighed less. Therefore, the above weighing scheme shifts the classifier's boundary closer to the negative points that are farther (in other words have smaller weights) while maintaining the distances of the closer ones and the positive class mostly unchanged. Refer to Algorithm \ref{appendix:maalgo} for a pseudo-code description.

\begin{figure}[t]
  \begin{minipage}[t]{.5\linewidth}
    \centering
    \begin{algorithm}[H]
    \KwData{$\mathbf{X}$, $\mathbf{Y}$, $\mathbf{C}$, g, $\lambda$}
    \KwResult{$\mathbf{w}$, $b$ and $\mathbf{Y}^p$}
    \tcc{Initialization by Vanilla SVM}
    $\lambda_{init} = 0$ ; $\mathbf{Y}^p$ = $\mathbf{Y}$\;
    $\mathbf{\mu}=$ QuadForm($\mathbf{X}$, $\mathbf{Y}$, $\mathbf{C}$, g, $\mathbf{Y}^p$, $\lambda_{init}$)\;
    Update $\mathbf{Y}^p$ = h($\mathbf{X}$)\;
    \tcc{SVM with a Recourse Constraint}
    \For {$\mathbf{Y} != \mathbf{Y}^p$}{
    $\mu$ $=$ QuadForm($\mathbf{X}, \mathbf{Y}, \mathbf{C}, g, \mathbf{Y}^p$, $\lambda$)\;
    Update $\mathbf{Y}^p$ = h($\mathbf{X}$)\;
    }
    \Return $h$
  \end{algorithm}
  \subcaption{Recourse regularized SVM classification}\label{algo:PPA}
  \end{minipage}
  \begin{minipage}[t]{.5\linewidth}
    \begin{algorithm}[H]
    \SetAlgoNoLine
    \KwData{$h$, $sw$, $n$}
    \KwResult{$h*$}
    \tcc{Choose best LIME fit}
    Choose best LIME samples $L^*$ in terms of prediction accuracy.\;
     $sw_0 \gets  1$\;
    Train classifier ($h$) with samples weights ($sw_0$)\;
    Find average recourse against $h$ using $L^*$\;
    Update sample weights ($sw$) using Eq.\ref{eqn:weightcalculation}\;
    Train classifier ($h^*$) with sample weights ($sw$)\;
    Find average recourse against $h^*$ using $L^*$\;
    \Return $h^*$
  \end{algorithm}
  \subcaption{Recourse equalization in black-box classification}
\label{appendix:maalgo}

  \end{minipage}
  \caption{Algorithms for Recourse Equalization a) Regularised SVM and b) Black-Box Classification}
  \label{fig:algos}
  \vspace{-1.0em}
\end{figure}
\section{Experiments}
\label{sec:experiments}
In this section, we will provide the details of our datasets and the experimental settings.

\textbf{Dataset description.} We evaluate both of our approaches on 4 datasets,
\credit \cite{yeh2009comparisons}, \german \cite{bache2013uci}, \givemecredit and \propublica.\footnote{\german: \url{https://github.com/algofairness/fairness-comparison/tree/master/fairness/data/preprocessed}, \givemecredit: \url{https://www.kaggle.com/c/GiveMeSomeCredit}, \propublica: \url{https://github.com/propublica/compas-analysis}}. We set the target variable as binary with labels +1 and -1 for all datasets. Positive sensitive attribute for \credit was chosen to be \emph{Married} and everything else as negative, the target value is -1 if the person defaults on a future credit card payment (we use a preprocessed version from \cite{ustun2019actionable} that uses the financial features). In the case of \givemecredit, after removing rows with missing information, the negative sensitive class was individuals with age \emph{under 35} and target is if the individual will be in financial distress in the next two years. We used \emph{gender} as sensitive attribute and set female as the negative class for both \german and \propublica. The target attribute for \german is credit risk of a person and for \propublica it is whether an individual would recommit a crime within 2 years from release. 

\vspace{-0.8em}
\paragraph{Experimental settings.} For every run, we choose a random sample from a dataset. We choose a random sample out of all 4 datasets which is $5,000$ datapoints for \credit (13 features), and $1,000$ for \givemecredit (10 features), and \propublica (402 features) and \german (59 features). We then use a $80/20$ split for train and test respectively. We perform 10-fold cross-validation in terms of minimizing recourse difference to choose the best parameter set for both linear and non-linear kernels. We search over $\lambda = \{0.2, 0.5, 1, 2, 10, 50, 100\}$ and polynomial kernel with $degree = \{2, 3, 5\}$. We then perform 10 iterations of our SVM-based algorithm using the optimum parameter set to obtain the final classifier. The costs for all the features are set to 1, and $\upsilon$ (cost parameter) is set to 10. We use the CVXOPT package \cite{andersen2013cvxopt} for solving the quadratic formulation, shown in \ref{svm:quadform}. 

In the agnostic setting, we run our experiments with $1000$ datapoints for all datasets using the following parameters. We execute 5 runs to choose 2 best set of samples for approximating black box classifier by LIME in terms of accuracy. We execute 10 independent runs of algorithm \ref{appendix:maalgo} and aggregate the results to generate the plots for all 4 datasets. We present our results on three black-box classifiers, namely Random Forest, Logistic Regression, and AdaBoost from the scikit-learn (v0.21) package \cite{scikit-learn}.\footnote{Specifically, \texttt{sklearn.ensemble.RandomForestClassifier}, \texttt{sklearn.linear\_model. LogisticRegression},  \texttt{sklearn.ensemble.AdaBoostClassifier}} We choose both Adaboost and Random forest primarily because they are non-linear in nature and lack the closed-form mathematical objective function that our previous method was based on. Our reason for choosing Logistic classifier is mainly to check how the black-box strategy works for simple linear classifiers. We use the Logistic and Adaboost classifier with default parameters and RandomeForest is trained with \emph{max-depth} (tree depth) set to 4 to avoid over-fitting and everything else is set as default.

In the SVM and model agnostic settings, we evaluate our proposed approaches for
equalizing recourse \emph{before} and \emph{after} regularization and sample
re-weighting, respectively. All results are reported with box and whisker plots
for complete details on the mean, median, and the percentiles of the accuracy
and the recourse difference distributions. The outliers are not shown in the
plot; however, they are considered in the computations.

\begin{figure}[!htb]
\minipage{0.25\textwidth}
  \includegraphics[width=\linewidth]{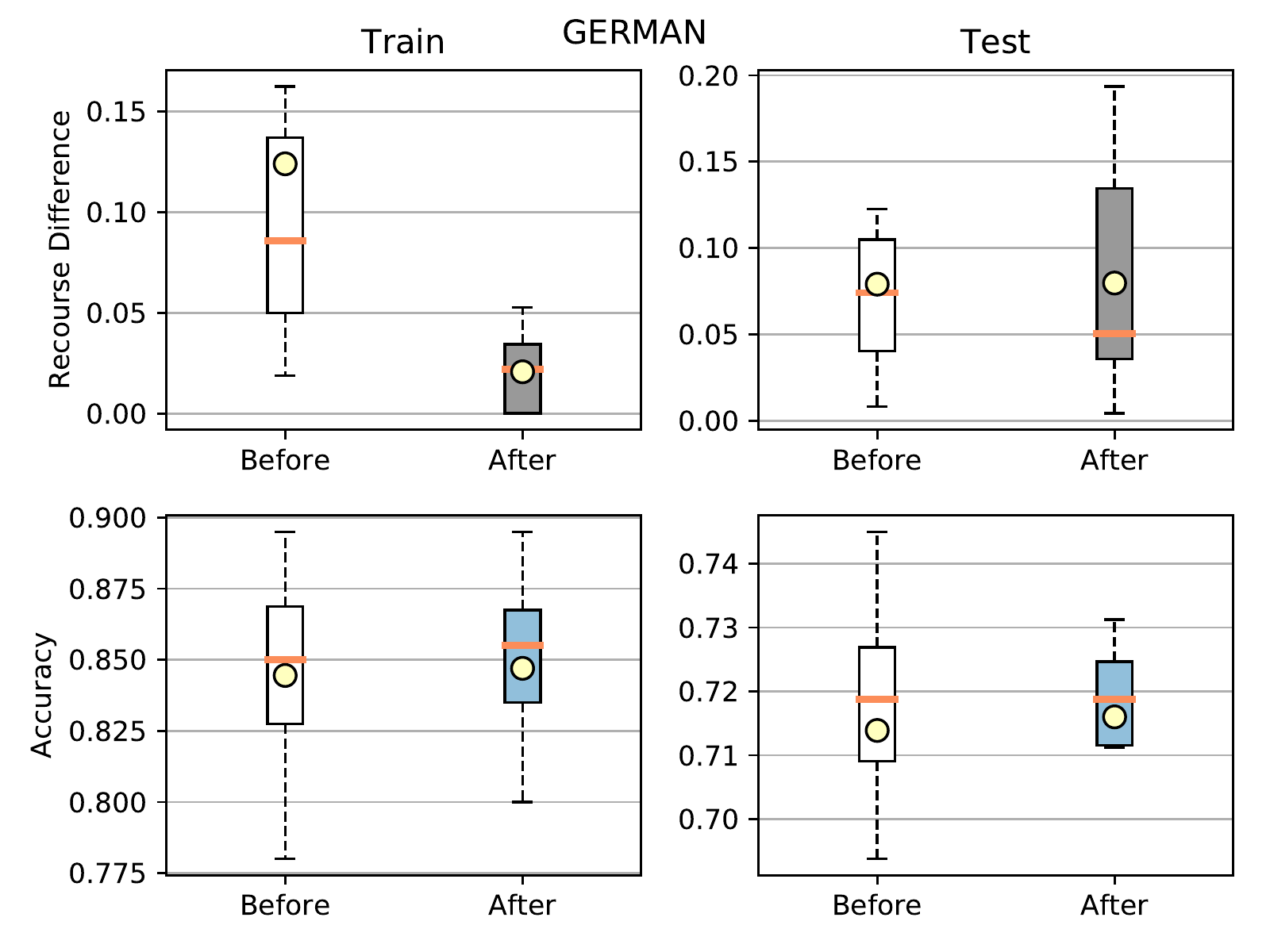}
\endminipage\hfill
\minipage{0.25\textwidth}
  \includegraphics[width=\linewidth]{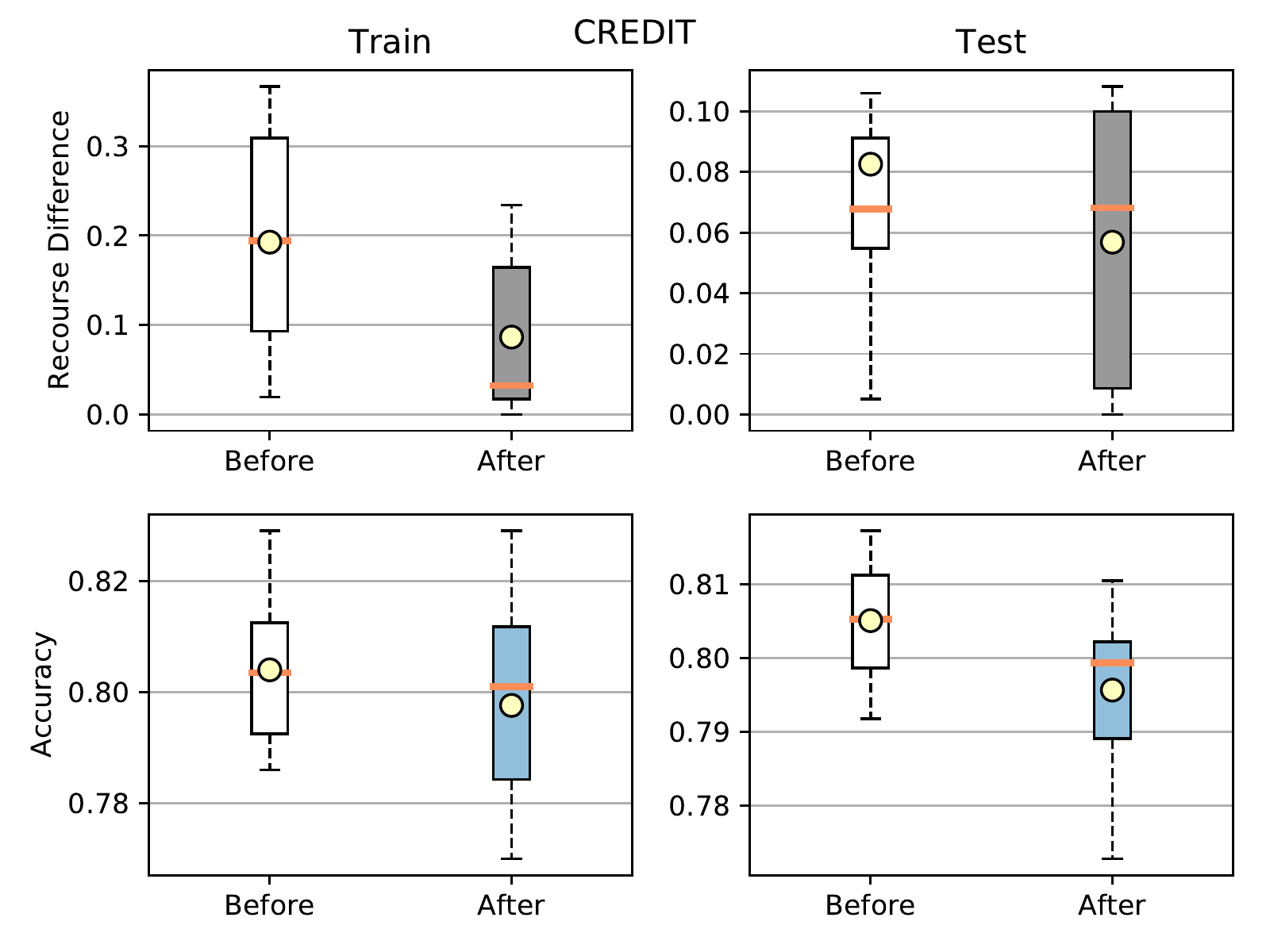}
\endminipage\hfill
\minipage{0.25\textwidth}%
  \includegraphics[width=\linewidth]{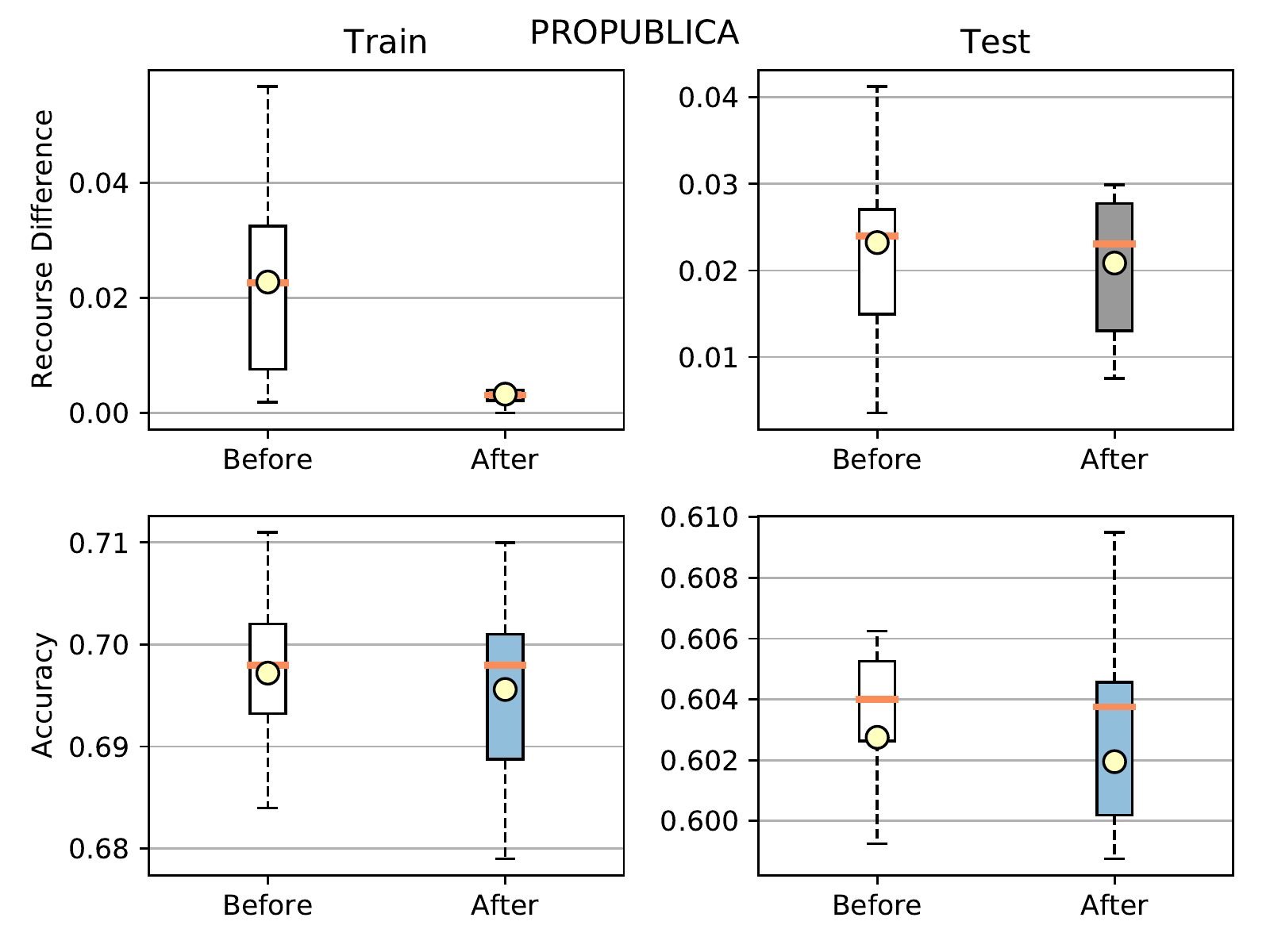}
\endminipage
\minipage{0.25\textwidth}%
  \includegraphics[width=\linewidth]{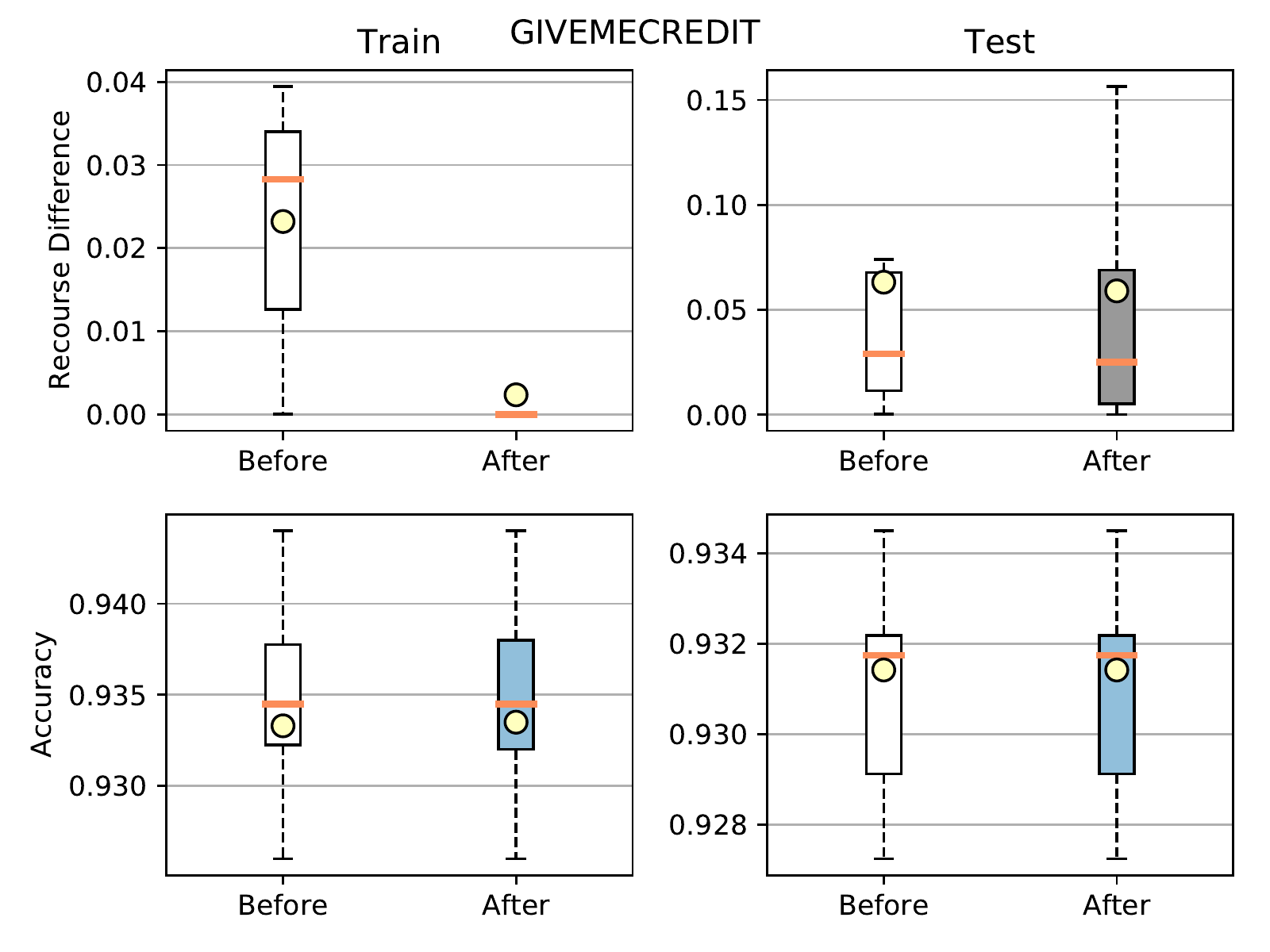}
\endminipage

\minipage{0.25\textwidth}
  \includegraphics[width=\linewidth]{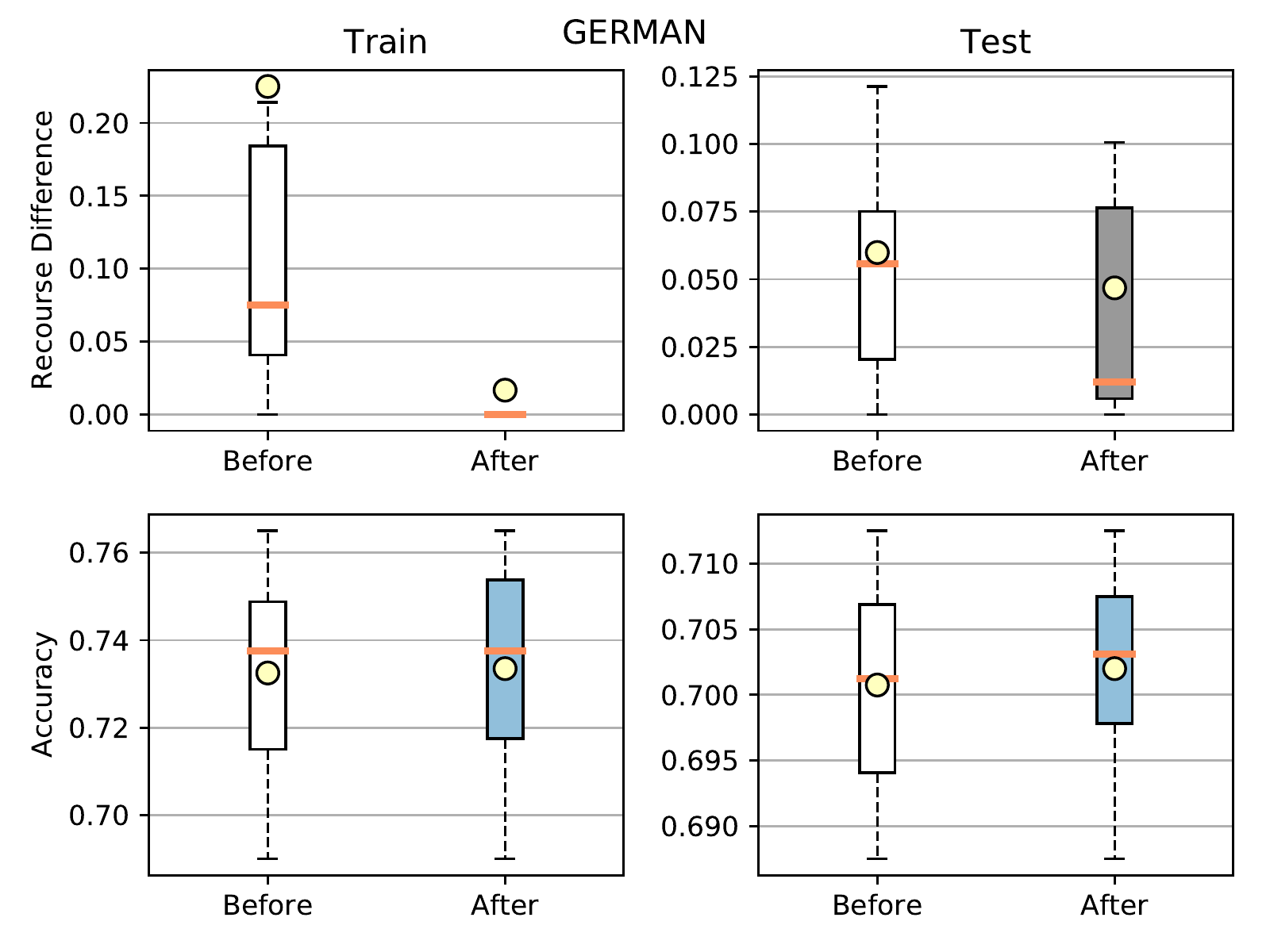}
\endminipage\hfill
\minipage{0.25\textwidth}
  \includegraphics[width=\linewidth]{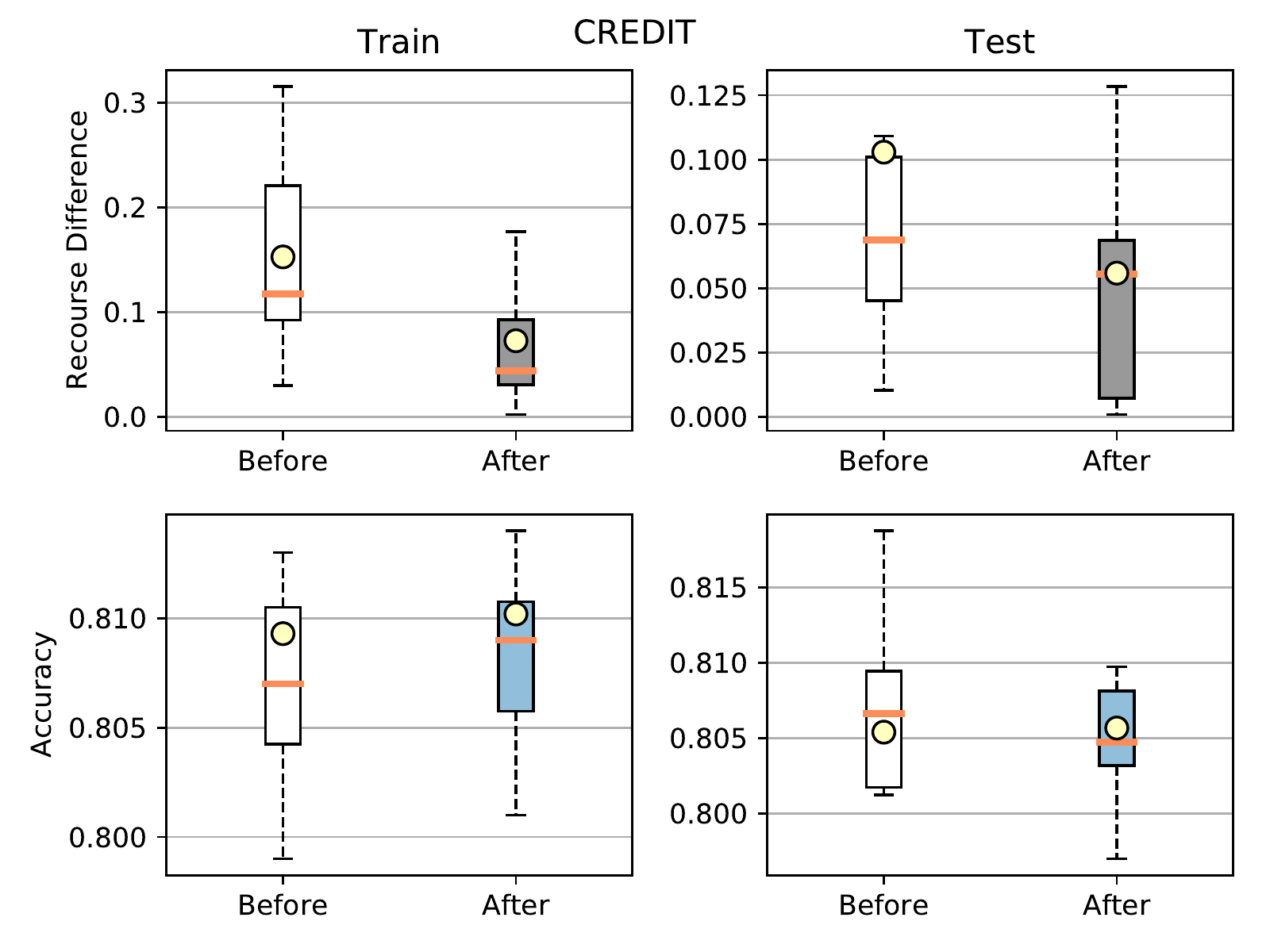}
\endminipage\hfill
\minipage{0.25\textwidth}%
  \includegraphics[width=\linewidth]{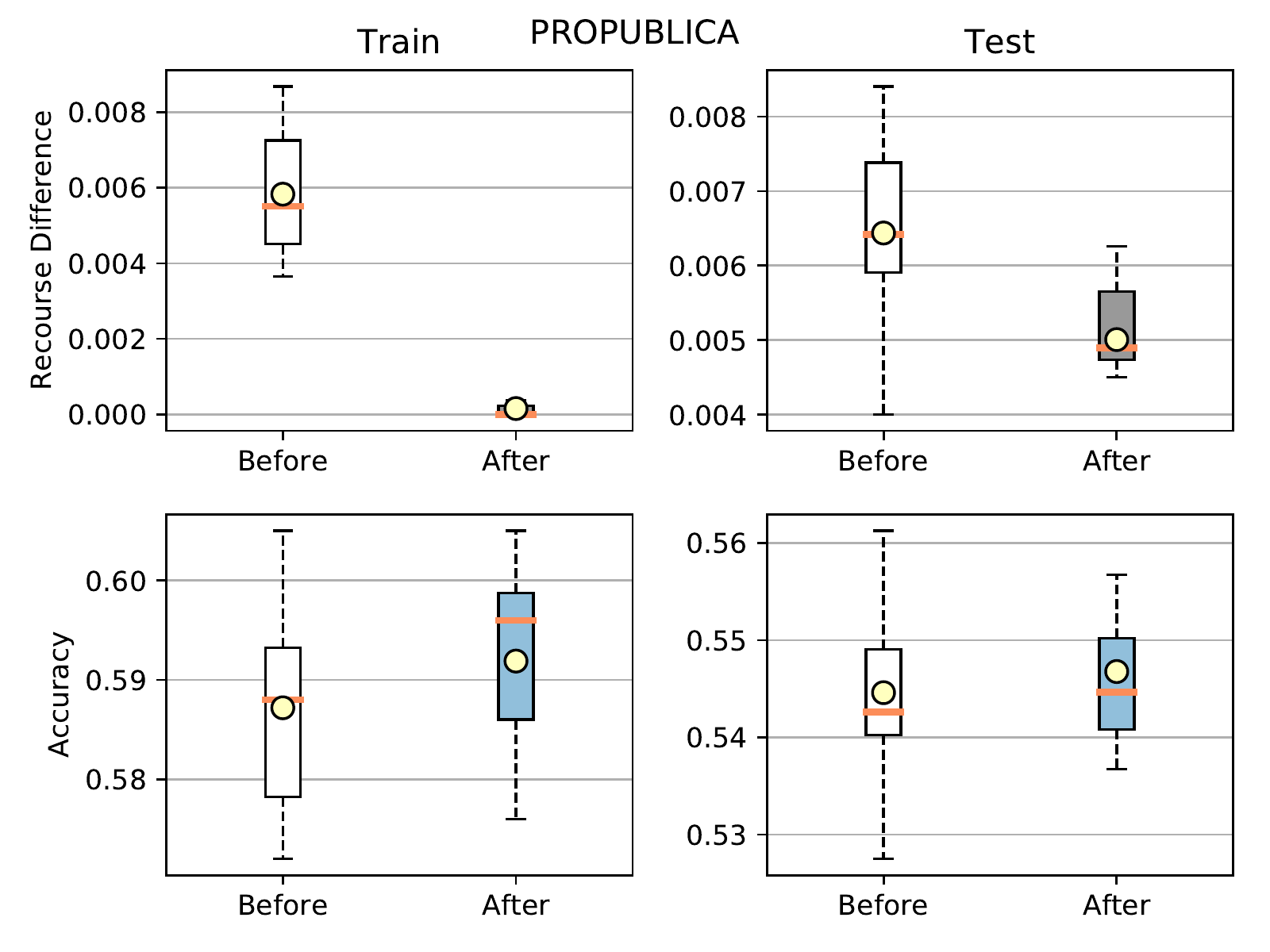}
\endminipage
\minipage{0.25\textwidth}%
  \includegraphics[width=\linewidth]{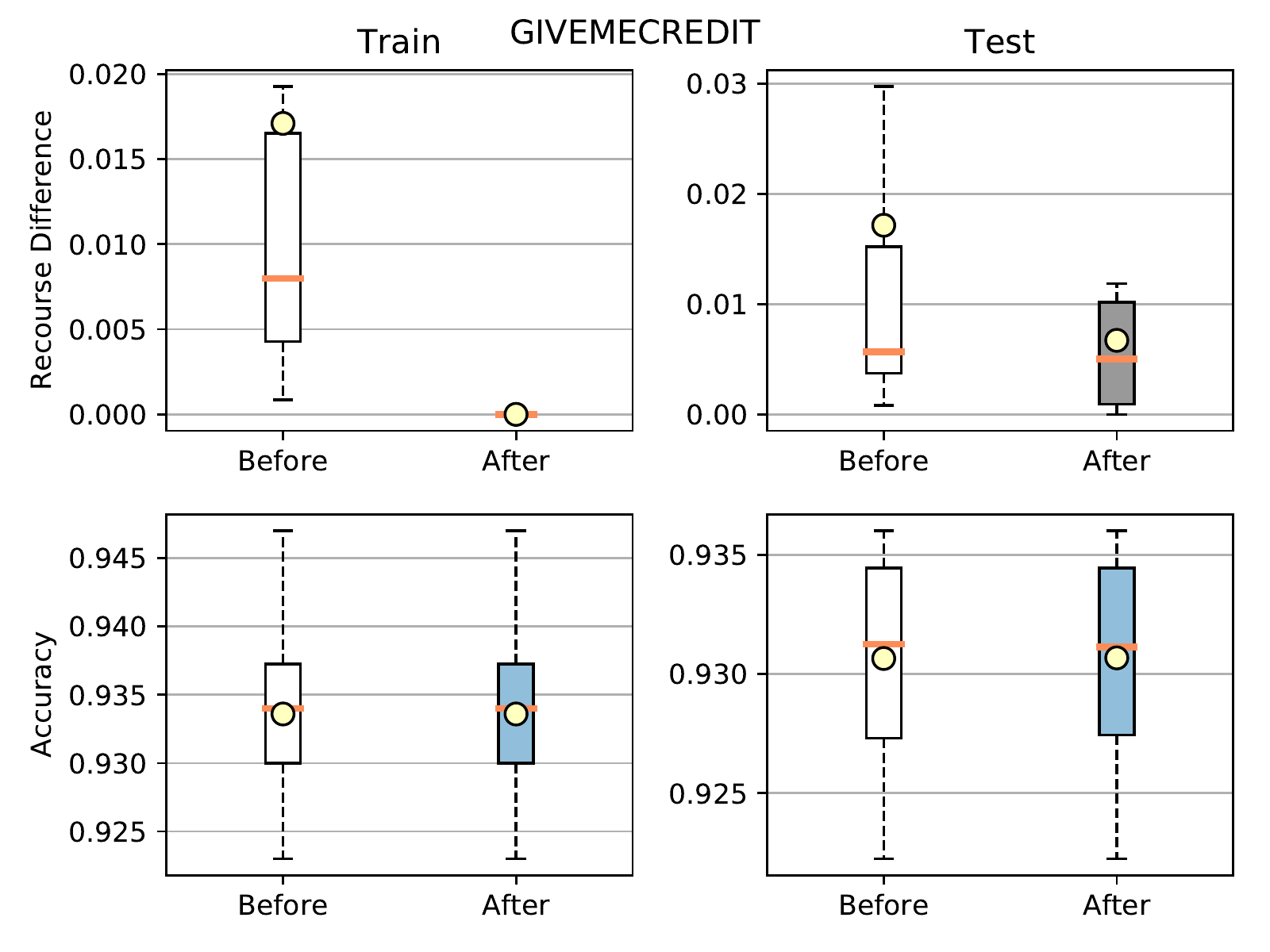}
\endminipage

 \caption{Linear (upper) and polynomial (lower) kernels' results on recourse and accuracy. Yellow circle is the mean, and the orange line is the median.}
   \vspace{-1.0em}
\label{fig:svm}
\end{figure}

\subsection{SVM results} We start by discussing the results of applying
Algorithm~\ref{algo:PPA}, listed in Table \ref{tab:svm}. We can summarize
these results by showing the percentage reduction in recourse difference ($\frac{\text{recourse\,\, before} - \text{recourse\,\,after}}{\text{recourse\,\,before}}$) for
different data sets (mean of the distributions). For this summary, we focus on the polynomial and linear kernels.

\begin{table}[htbp]
  \centering
    \caption{Summary of percentage reduction in recourse difference for polynomial and linear kernels}
    \medskip
   \begin{tabular}{r|cc|cc}\toprule
   & \multicolumn{2}{c|}{Linear} & \multicolumn{2}{c}{Polynomial}\\ \hline
    Dataset & Train & Test & Train & Test \\  \midrule
    \german  &80\% & 0\% & 91\% & 23\%\\
    \credit  & 52\% & 29\% & 51\%&  50\% \\
    \propublica  &84\% & 16\% & 99\% & 23\%\\
    \givemecredit & 90\%& 9\%& 100\%& 55\% \\
    \bottomrule
  \end{tabular}
  \label{tab:svm}
    \vspace{-1.0em}
\end{table}

In general, we observe that we are yielding an improvement in the unfairness of recourse across groups in terms of the means of the difference. We also observe that the medians are
decreasing (and we have smaller ranges of percentile distributions after applying our algorithm), which means at least around half of the runs are yielding a smaller recourse difference compared to \emph{before} even though the mean might remain the same due to outliers, e.g., in \german the test recourse difference is not decreasing significantly `after' applying our algorithm compared to other datasets. However, more than $50\%$ (e.g., lower $50$th percentile for both kernels) of the runs have a smaller recourse difference \emph{after} according to the median, or up to $75$th percentile of \givemecredit data have a smaller recourse. In addition, we observe that the accuracy roughly stays the same (it decreases no more than 1\% and sometimes even slightly increases). Furthermore, the improvement we acquired using the linear kernel is less compared to polynomial, we assume this is due to fewer degrees of freedom (fewer model parameters) in linear kernel compared to polynomial to change the decision boundary. 

\begin{figure}[!htbp]
\minipage{0.32\textwidth}
  \includegraphics[width=\linewidth]{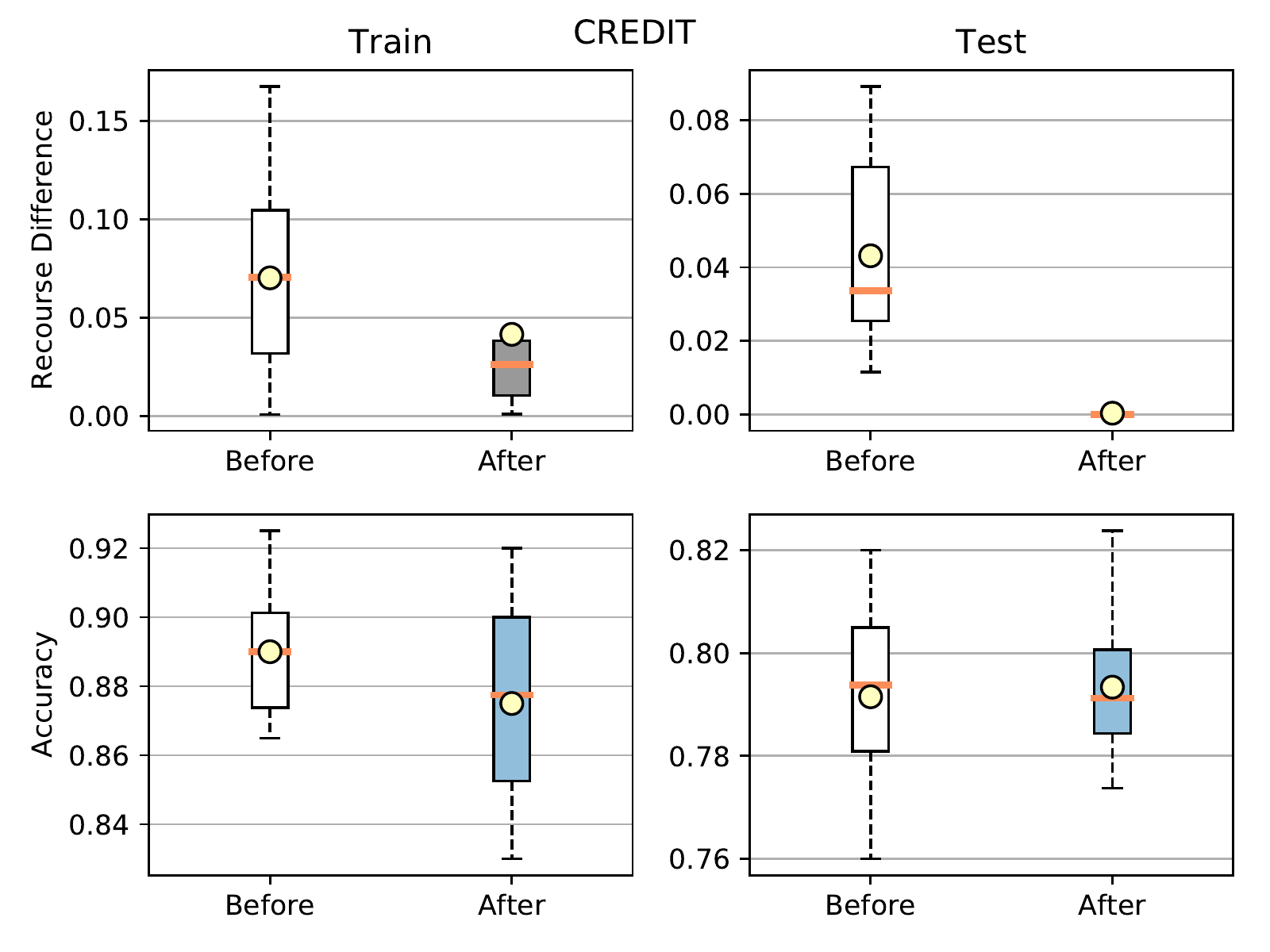}
\endminipage\hfill
\minipage{0.32\textwidth}
  \includegraphics[width=\linewidth]{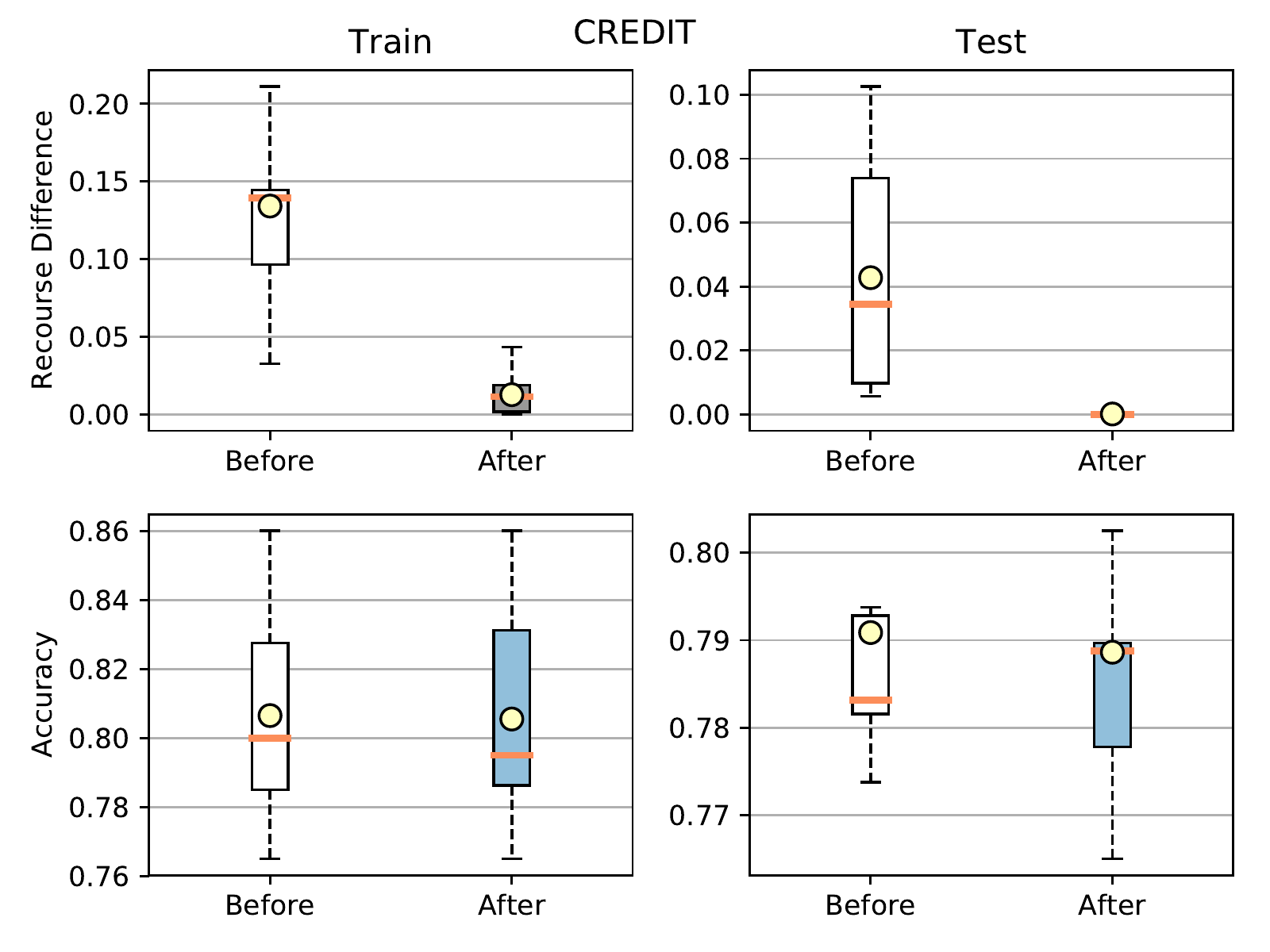}
\endminipage\hfill
\minipage{0.32\textwidth}%
  \includegraphics[width=\linewidth]{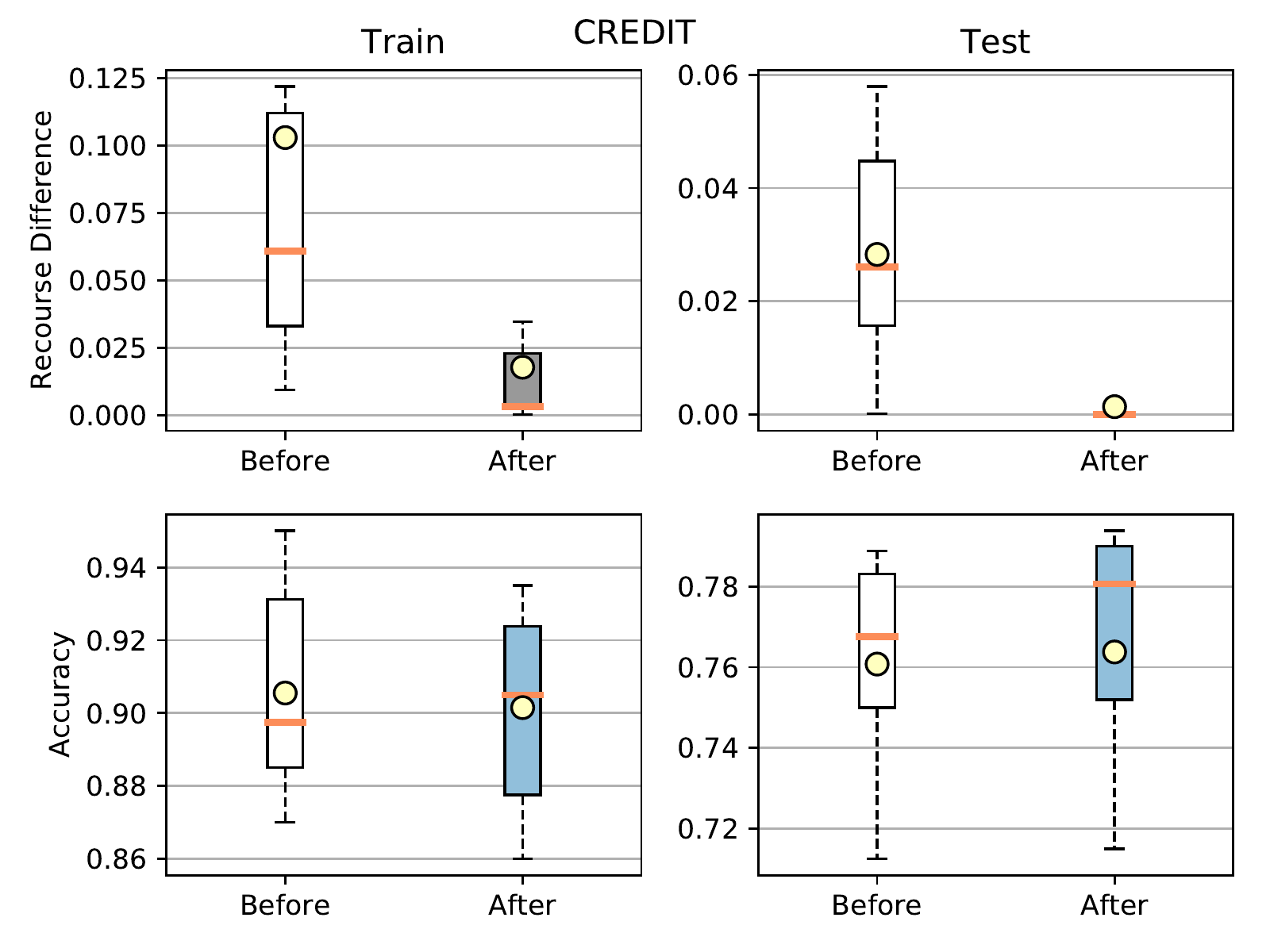}
\endminipage

\minipage{0.32\textwidth}
  \includegraphics[width=\linewidth]{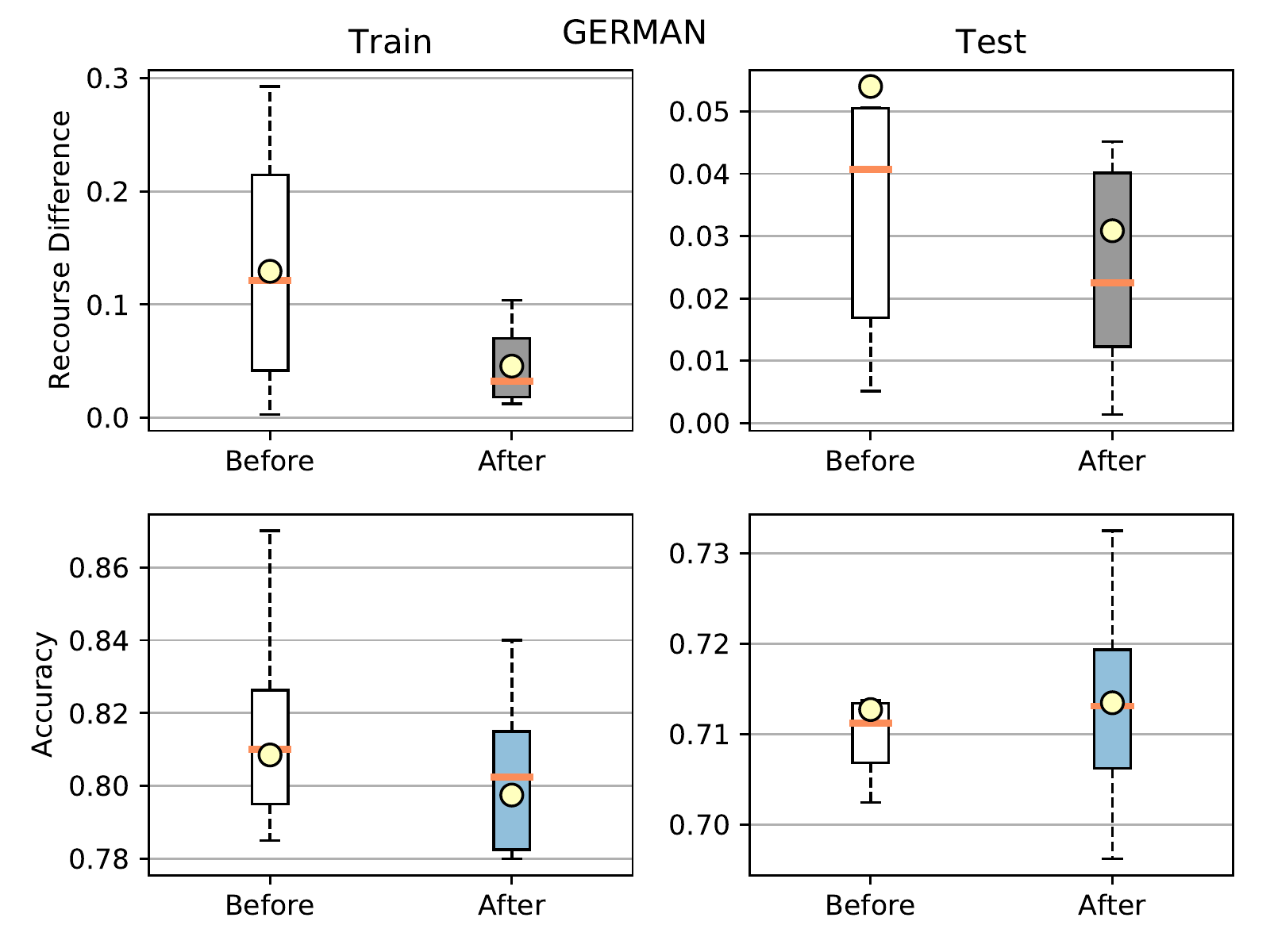}
\endminipage\hfill
\minipage{0.32\textwidth}
  \includegraphics[width=\linewidth]{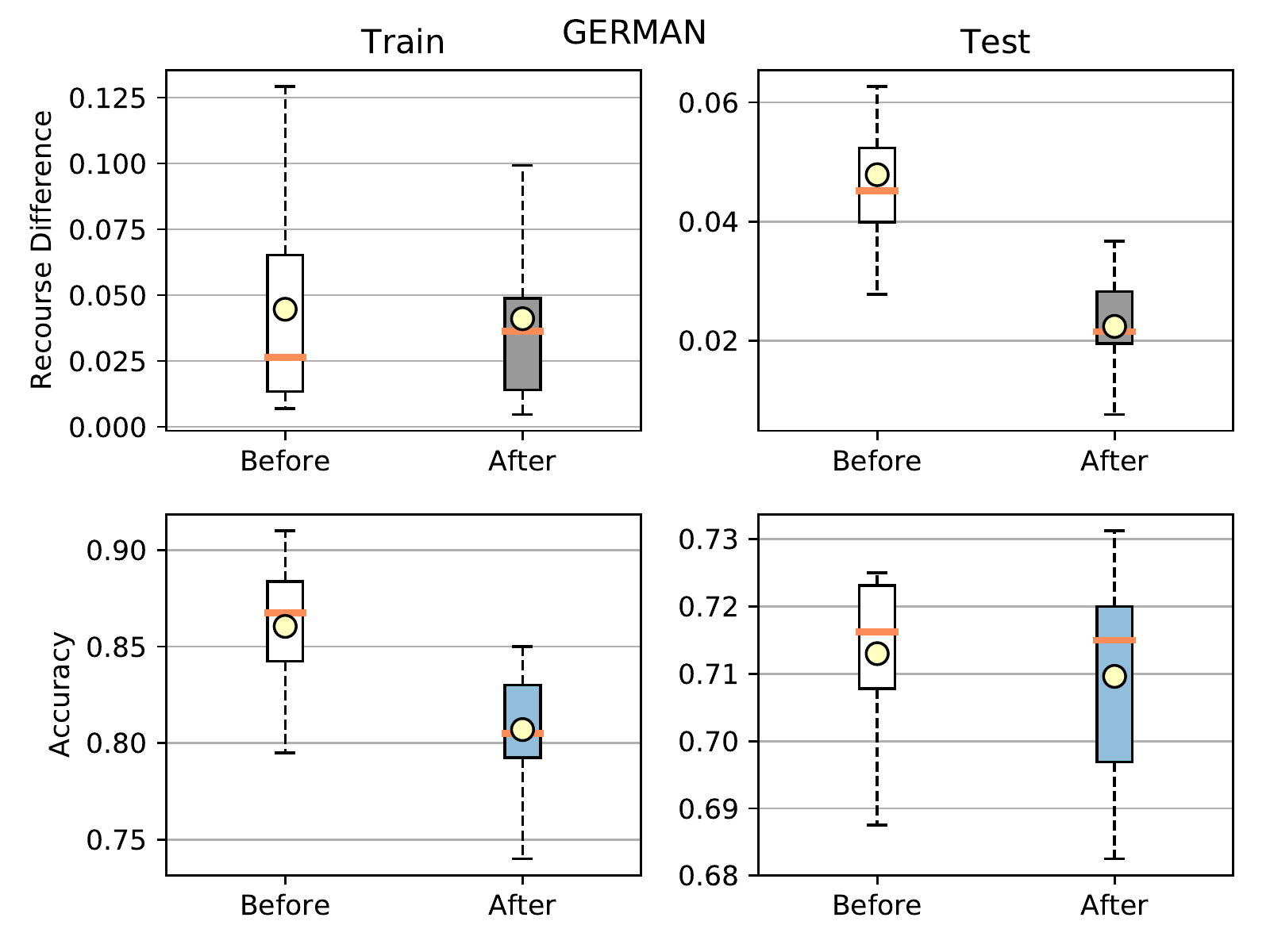}
\endminipage\hfill
\minipage{0.32\textwidth}%
  \includegraphics[width=\linewidth]{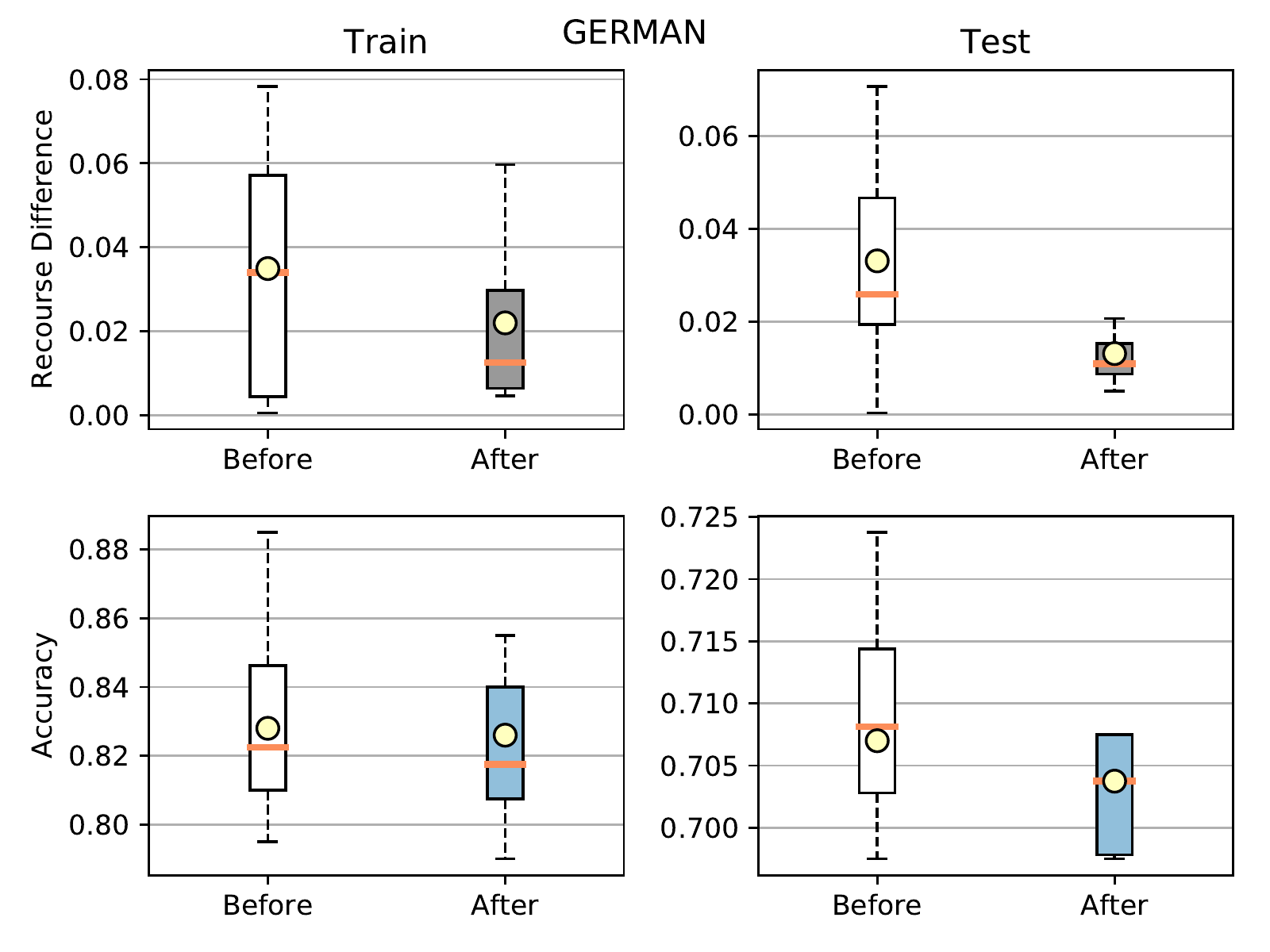}
\endminipage

\minipage{0.32\textwidth}
  \includegraphics[width=\linewidth]{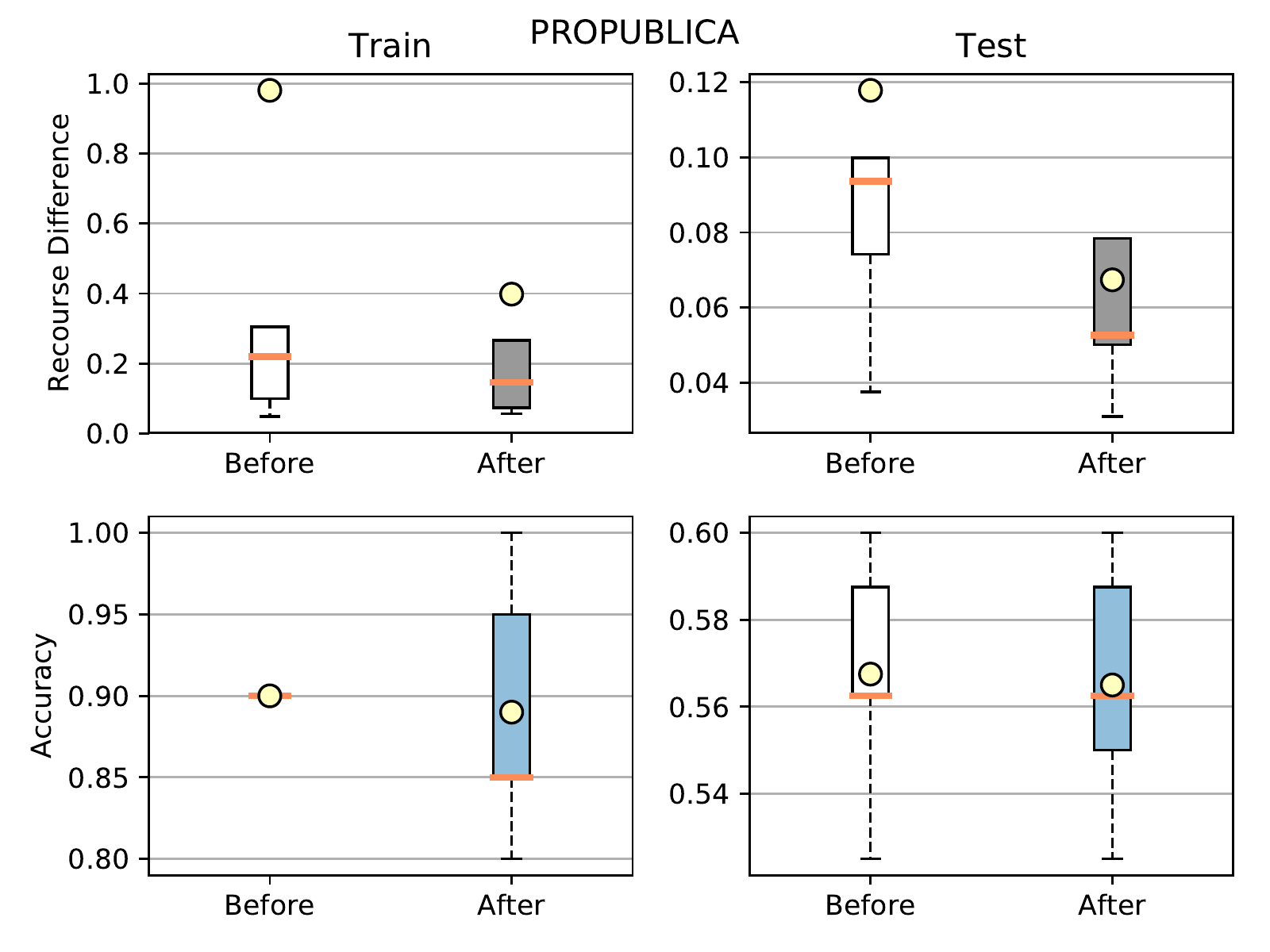}
\endminipage\hfill
\minipage{0.32\textwidth}
  \includegraphics[width=\linewidth]{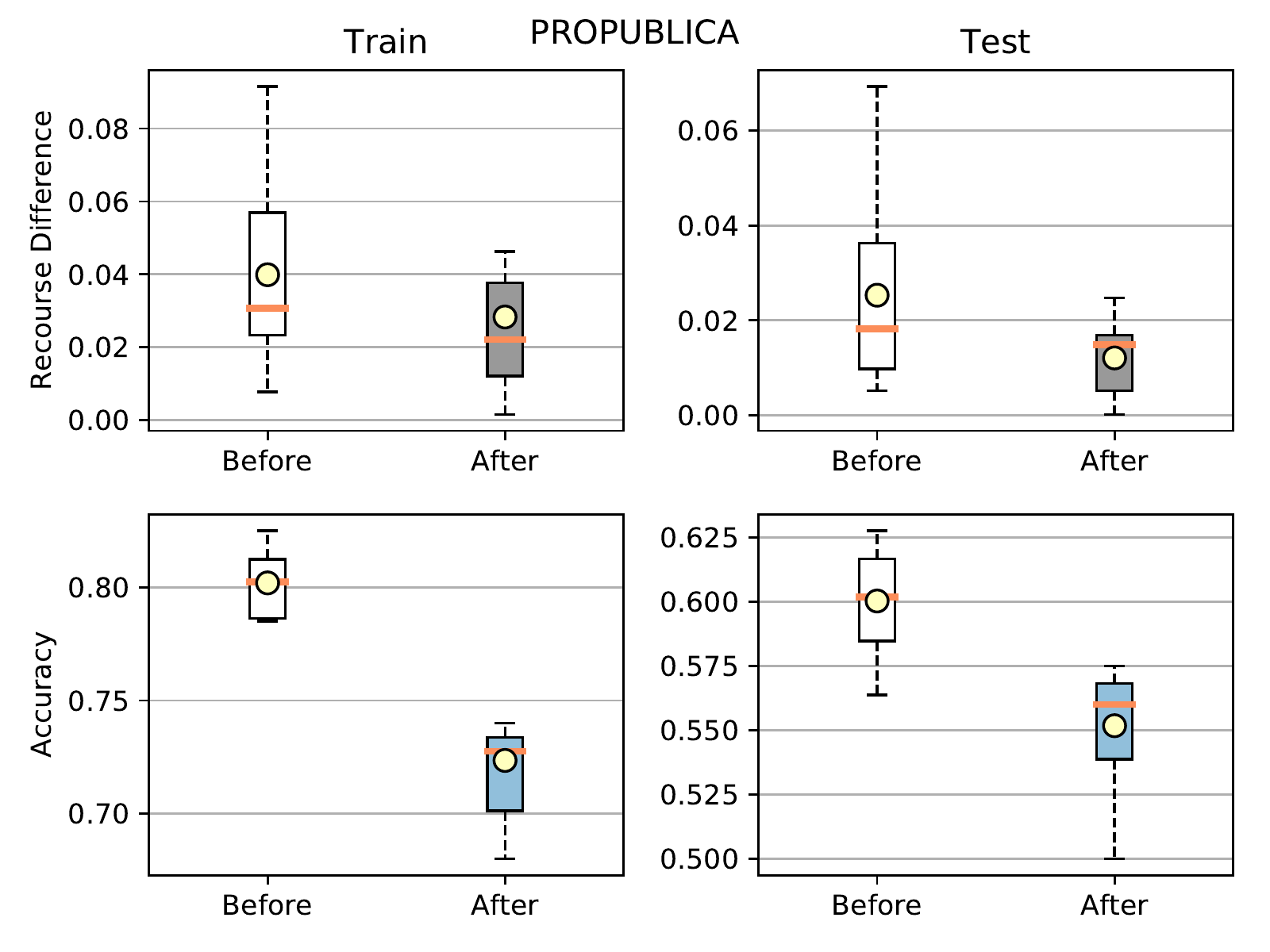}
\endminipage\hfill
\minipage{0.32\textwidth}%
  \includegraphics[width=\linewidth]{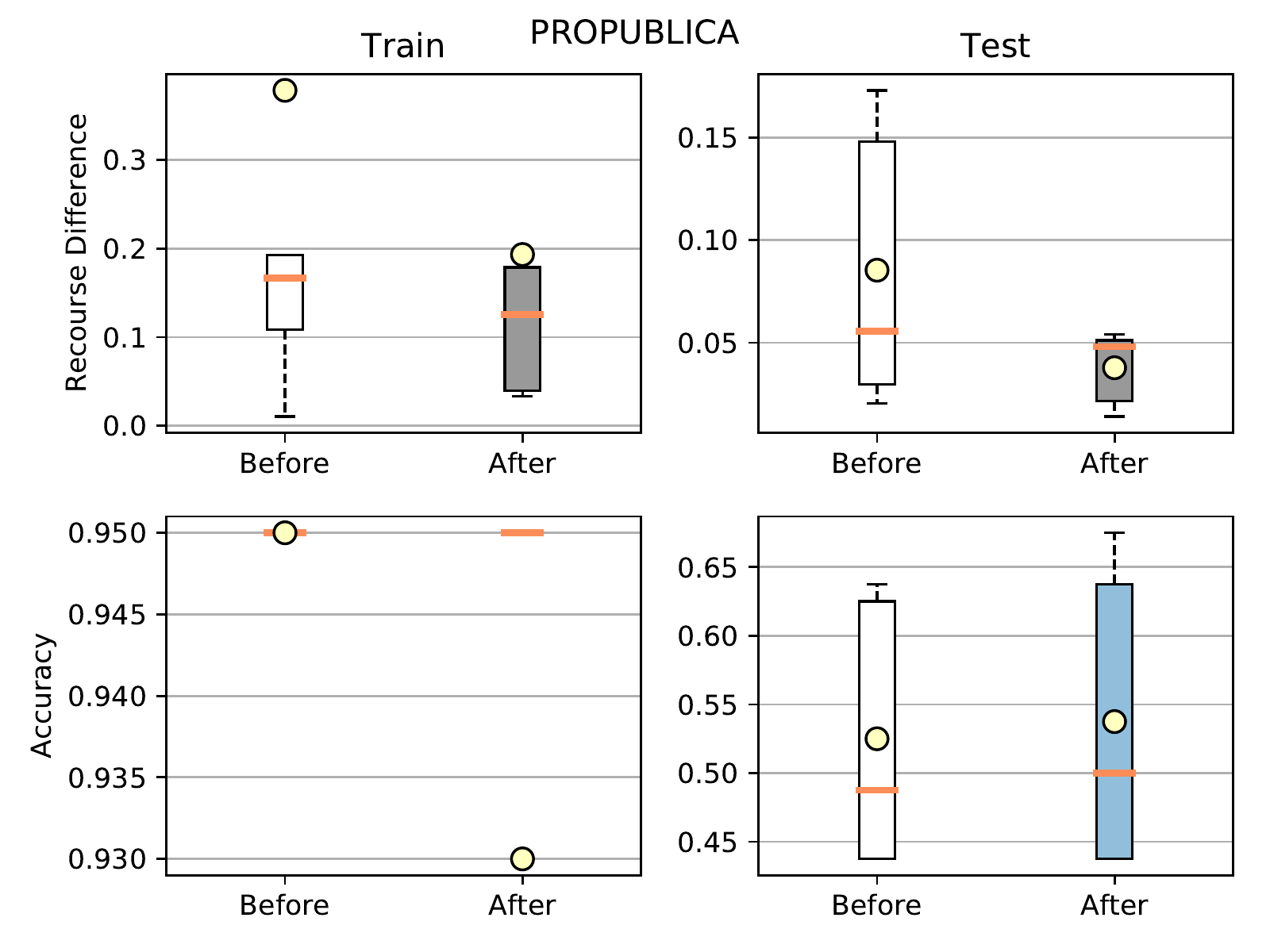}
\endminipage

\minipage{0.32\textwidth}
  \includegraphics[width=\linewidth]{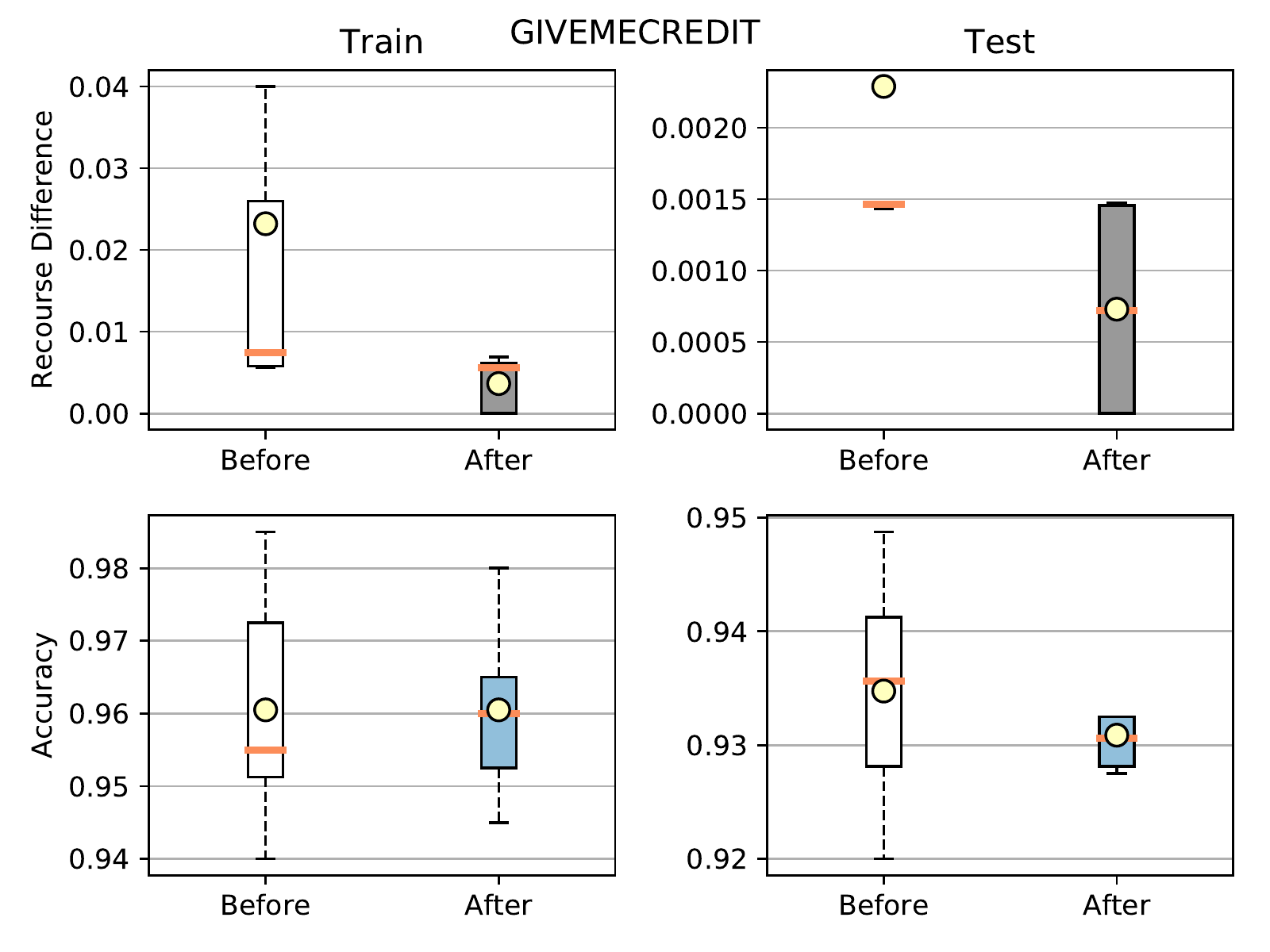}
\endminipage\hfill
\minipage{0.32\textwidth}
  \includegraphics[width=\linewidth]{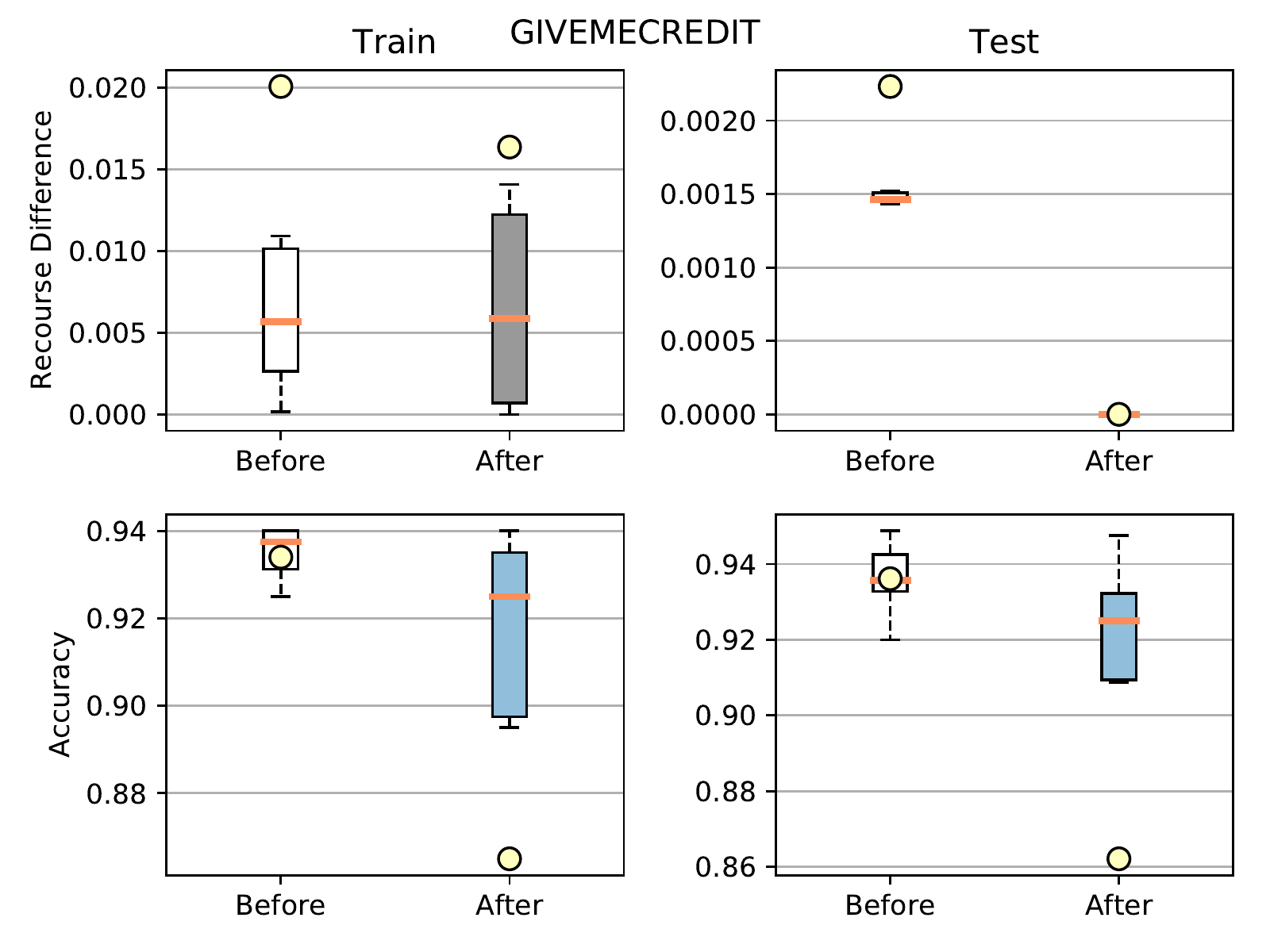}
\endminipage\hfill
\minipage{0.32\textwidth}%
  \includegraphics[width=\linewidth]{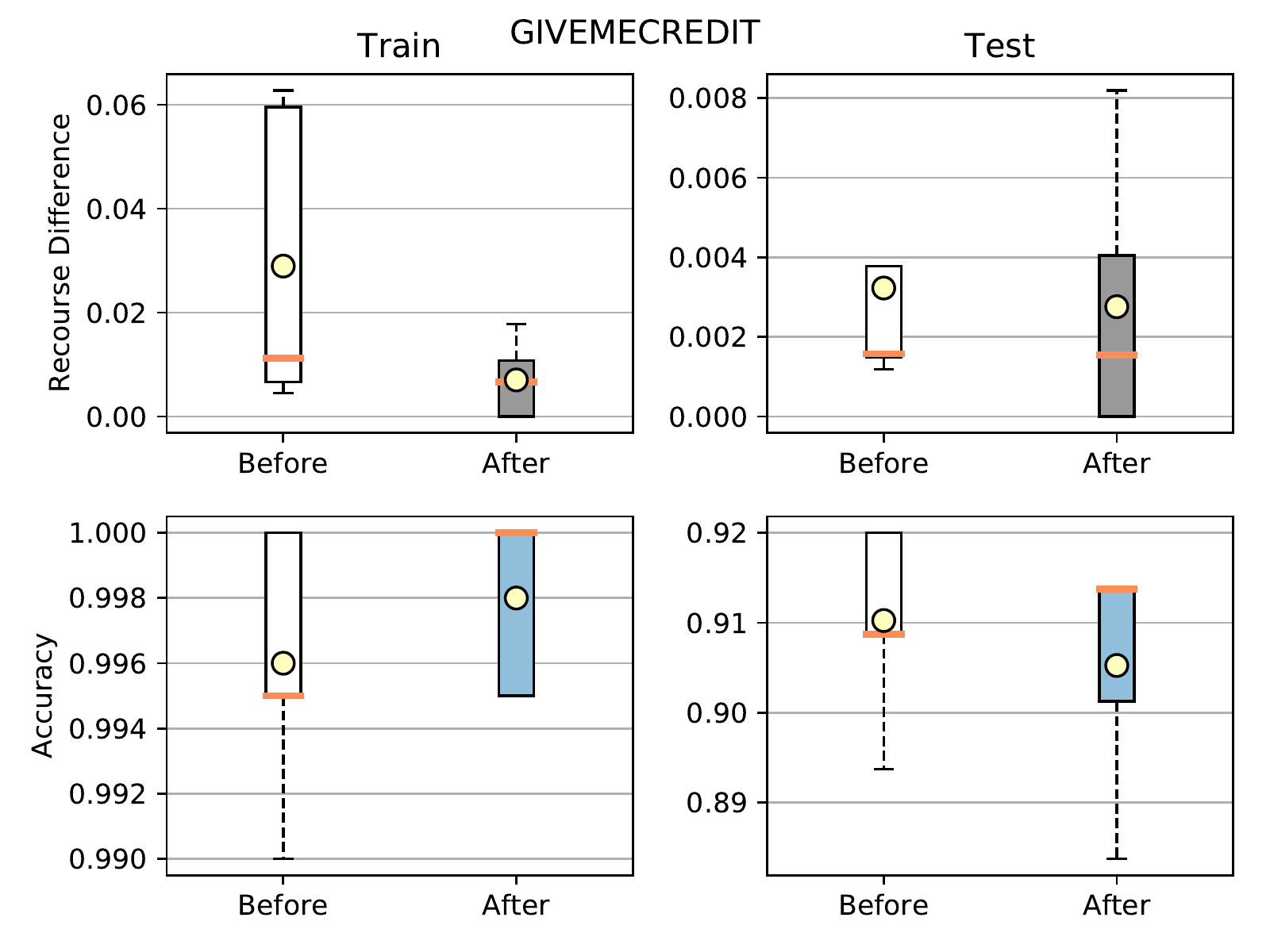}
\endminipage

\caption{RandomForest (left), LogisticRegression (middle), AdaBoost (right) classifier's results on recourse and accuracy. Yellow circle is the mean and the orange line is the median.}
\label{fig:agnostic}
\end{figure}

\subsection{Model agnostic results} 
We have listed the results of applying Algorithm \ref{appendix:maalgo} in Table \ref{tab:agno}. Similar to SVM, both train and test sets show improvements in terms of the mean, median and percentile distributions of the recourse difference \emph{before} and \emph{after} equalization. In addition, the mean accuracies for \credit, \german, and \propublica are not reduced by more than 2\%, except for the training accuracy during logistic regression for \german ($6$\% drop), test/train accuracies for \propublica (around 7\% and 5\% drops, respectively), and test/train accuracies for \givemecredit (around 7\% for both). 

\textbf{General Analysis.} Overall, these results show that our proposition to define recourse for a group as the mean of the distances from the decision boundary and its incorporation in the SVM formulation as a regularizer can reduce the unfairness in the recourse across groups. Additionally, the simple re-weighting strategy works well in terms of recourse difference reduction for classifiers that support sample weights and could be useful for black-box scenarios where more fine-grained control cannot be obtained for explicit distances from the boundary in terms of model parameters. Both methods yield these results without affecting the accuracy significantly. In addition, the median and percentage-based distributions of the recourse difference across various runs decrease after equalization.

\begin{table}[htbp]
  \vspace{-1.0em}
  \caption{Summary of percentage reduction in recourse difference for Model Agnostic}
  \medskip
  \centering
  \begin{tabular}{r|cc|cc|cc}\toprule
   & \multicolumn{2}{c|}{Random Forest} & \multicolumn{2}{c|}{Logistic Regression} & \multicolumn{2}{c}{AdaBoost}\\ \hline
    Dataset & Train & Test & Train & Test & Train & Test \\  \midrule
    \credit  & 41\% & 99\% & 90\% & 100\% & 83\% & 95\%\\
    \german  & 64\% & 42\% & 8\% & 53\% & 37\% & 60\%\\
    \propublica  & 59\% & 43\%&  29\% & 52\% & 49\% & 56\%\\
    \givemecredit & 83\% & 70\% & 20\% & 100\% & 66\% & 14\%\\
    \hline
  \end{tabular}
  \label{tab:agno}
  \vspace{-1.0em}
\end{table}

\section{Conclusion and future work}
\label{sec:conclusion}

  In summary, this work introduces a new notion of fairness in recourse (Equalizing Recourse), i.e., classifiers that maintain a good performance while providing opportunities for feasible recourse across groups. We proposed various approaches based on a general definition of recourse as the absolute value of the difference between the average of the distance of all points from the boundary in a group. We then utilized this definition to make classification models fair in terms of recourse. Our approach is generalizable to linear and non-linear settings, including kernelizable (support vector machines) and non-kernelizable (decision trees) models. We empirically validated these methods through intensive experiments. This is the first work on outcome-independent fairness in terms of recourse (equal costs for changing the outcome) as far as our knowledge.  Incorporating multiple groups/sensitive attributes is a possible future direction. In addition, instead of only equalizing the means of the distributions, we can equalize the distributions themselves using correlation or KL divergence. 

\bibliographystyle{alpha}  
\bibliography{references}

\end{document}